\documentclass[times,twocolumn,final]{elsarticle}

\usepackage{medima}
\usepackage{framed,multirow}

\usepackage{amssymb}
\usepackage{amsmath}
\usepackage{latexsym}

\usepackage{url}
\usepackage{xcolor}
\usepackage{pifont} 
\usepackage{comment}

\newcommand{\cmark}{\ding{51}}
\newcommand{\xmark}{\ding{55}}
\newcommand{\specialcell}[2][c]{%
  \begin{tabular}[#1]{@{}l@{}}#2\end{tabular}}

\usepackage{hyperref}

\definecolor{newcolor}{rgb}{.8,.349,.1}

\def\colorrevone{black}
\def\colorrevtwo{black}

\journal{Medical Image Analysis}

\begin{document}

\verso{Dominik Rivoir \textit{et~al.}}

\begin{frontmatter}

\title{On the Pitfalls of Batch Normalization for End-to-End Video Learning: A Study on Surgical Workflow Analysis}

\author[1,2]{Dominik \snm{Rivoir}\corref{cor1}}
\cortext[cor1]{Corresponding author at: 
  Department of Translational Surgical Oncology, National Center for Tumor Diseases (NCT/UCC), Fetscherstraße 74, 01307 Dresden, Germany}
\ead{dominik.rivoir@nct-dresden.de}
\author[1,2]{Isabel \snm{Funke}}
\author[1,2]{Stefanie \snm{Speidel}}

\address[1]{Department of Translational Surgical Oncology, National Center for Tumor Diseases (NCT/UCC Dresden), Fetscherstraße 74, 01307 Dresden, Germany: German Cancer Research Center (DKFZ), Heidelberg, Germany; Faculty of Medicine and University Hospital Carl Gustav Carus, TUD Dresden University of Technology, Dresden, Germany; Helmholtz-Zentrum Dresden-Rossendorf (HZDR), Dresden, Germany}
\address[2]{Centre for Tactile
Internet with Human-in-the-Loop (CeTI), TUD Dresden University of Technology, Dresden, Germany}

\received{Jul 31, 2023}
\accepted{Feb 26, 2024}

\begin{abstract}
Batch Normalization's (BN) unique property of depending on other samples in a batch is known to cause problems in several tasks, including sequence modeling. Yet, BN-related issues are hardly studied for long video understanding, despite the ubiquitous use of BN in CNNs (Convolutional Neural Networks) for feature extraction. Especially in surgical workflow analysis, where the lack of pretrained feature extractors has led to complex, multi-stage training pipelines, limited awareness of BN issues may have hidden the benefits of training CNNs and temporal models end to end. In this paper, we analyze pitfalls of BN in video learning, including issues specific to online tasks such as a 'cheating' effect in anticipation. We observe that BN's properties create major obstacles for end-to-end learning. However, using BN-free backbones, even simple CNN-LSTMs beat the state of the art {\color{\colorrevtwo}on three surgical workflow benchmarks} by utilizing adequate end-to-end training strategies which maximize temporal context. We conclude that awareness of BN's pitfalls is crucial for effective end-to-end learning in surgical tasks. By reproducing results on natural-video datasets, we hope our insights will benefit other areas of video learning as well. Code is available at: \url{https://gitlab.com/nct_tso_public/pitfalls_bn}
\end{abstract}


\end{frontmatter}



\section{Introduction}
Batch Normalization (BatchNorm/BN)~\citep{ioffe2015batch} is a highly effective regularizer in visual recognition tasks and is ubiquitous in modern Convolutional Neural Networks (CNNs)~\citep{he2016deep,szegedy2016rethinking,tan2019efficientnet}. It is, however, also considered a source for silent performance drops and bugs due to its unique property of depending on other samples in the batch and the assumptions tied to this~\citep{brock2021high,wu2018group,wu2021rethinking}. Most notably, BatchNorm  
assumes 
that batches are a good approximation of the training data and only performs well when batches are large enough and sampled i.i.d.~\citep{ioffe2015batch,wu2021rethinking}.

This generally does not hold for sequence learning, where batches contain highly correlated, sequential samples, and has led to the use of alternatives such as LayerNorm (LN)~\citep{ba2016layer} in NLP~\citep{shen2020powernorm,vaswani2017attention}. In video learning, BN has been studied less~\citep{cai2021dynamic,wu2018group}, despite the use of BN-based CNNs.

In natural-video tasks, CNNs are only used to extract image- or clip-wise features using pretrained CNNs (e.g.~\citep{carreira2017quo}) off-the-shelf. Only the temporal model, which typically does not contain BN, is trained to aggregate features over time~\citep{abu2020long,farha2019ms,huang2020improving,ishikawa2021alleviating,ke2019time,sener2020temporal,wang2020boundary,chinayi_ASformer}. However, in specialized small-data domains such as surgical video, well-pretrained CNNs may not be available~\citep{czempiel2022surgical,zhang2022large}, requiring CNNs to be finetuned, either through 2-stage~\citep{czempiel2020tecno} or end-to-end (E2E) training~\citep{jin2017sv}. The latter seems preferable to enable joint learning of visual and temporal features, especially since spatio-temporal feature extractors (e.g.\ 3D CNNs) have not been effective on small-scale surgical datasets~\citep{czempiel2022surgical,zhang2022large}. 
However, BN layers in CNNs pose obstacles for end-to-end learning.

We hypothesize that BN's problems with correlated, sequential samples have silently caused research in video-based surgical workflow analysis~(SWA) to head into a sub-optimal direction.
The focus has shifted towards developing sophisticated temporal models to operate on extracted image features similar to the natural-video domain, replacing end-to-end learning with complex multi-stage training procedures, where each component (CNN, LSTM~\citep{hochreiter1997long}, TCN~\citep{farha2019ms}, Transformer~\citep{vaswani2017attention}, etc.) is trained individually \citep{bano2020fetnet,gao2021trans,kannan2019future,marafioti2021catanet,yuan2021surgical}.
We argue that even simple CNN-LSTM models can often outperform these methods when BN-free backbones are used and the model is trained end to end.

We investigate BatchNorm's pitfalls in end-to-end learning on two surgical workflow analysis (SWA) tasks:
\emph{surgical phase recognition}~\citep{twinanda2016endonet} and \emph{instrument anticipation}~\citep{rivoir2020rethinking}. 
We choose these for two reasons:
The lack of well-pretrained feature extractors and ineffectiveness of 3D~CNNs signify the need for end-to-end approaches in SWA and thus make BN-issues most relevant here. Further, SWA is one of the most active research areas for \emph{online} video  understanding,
where models are constrained to access only past frames, causing additional BN problems regarding leakage of future information.

Our contributions can be summarized as follows:
\begin{enumerate}
    \item We challenge the predominance of multi-stage approaches in surgical workflow analysis (SWA) and show that awareness of BN's pitfalls is crucial for effective end-to-end learning.
    \item We provide a detailed literature review showing BN's impact on existing training strategies in surgical workflow analysis.
    \item We analyze when BN issues occur, including problems specific to online tasks such as ``cheating'' in anticipation.
    \item Leveraging these insights, we show that even simple, end-to-end CNN-LSTM models can be highly effective when BN is avoided and strategies which maximize temporal context are employed.
    \item These CNN-LSTMs beat the state of the art {\color{\colorrevtwo}on three surgical workflow benchmarks}.
    \item We reproduce our findings on natural-video datasets to show our study's potential for wider impact.
\end{enumerate}

\section{Background \& Related Work}

\subsection{Batch Normalization and Beyond}

BatchNorm~\citep{ioffe2015batch} is used in almost all modern CNNs. It is assumed to have a regularizing effect, improve convergence speed and enable training of deeper networks~\citep{bjorck2018understanding,de2020batch,hoffer2017train}.
In BatchNorm, features~$x$ are standardized by estimating the channel-wise mean and variance and then transformed using an affine function with learnable parameters $\gamma,\beta$. Formally, 
\begin{equation}
    BN(x) = \gamma \frac{x - E(x)}{\sqrt{Var(x) + \epsilon}} + \beta ,
\end{equation}
where $E(x)$ and $Var(x)$ are the mean and variance estimates calculated from the current batch during training.
During inference, when the concept of batches is generally not applicable anymore, a running average of the training values of $E(x)$ and $Var(x)$ is used to estimate the global dataset statistics.

Despite BN's popularity and success, it has major disadvantages. The discrepancy between train and test behavior can lead to drops in performance when batch statistics are poor estimates of global statistics, e.g.\ when batches are small or contain correlated samples. Also, BN's 
property of depending on other samples in the batch can lead to hidden mistakes if not handled with care. Resulting problems have been observed in many areas such as object detection~\citep{wu2018group,wu2021rethinking}, sequence modeling~\citep{ba2016layer}, reinforcement learning~\citep{salimans2016weight}, continual learning~\citep{pham2022continual}, federated learning~\citep{andreux2020siloed} and contrastive learning~\citep{henaff2020data}. E.g.\ in language modeling, both aforementioned issues are highly problematic and have led to the use of alternative normalization layers~\citep{ba2016layer,shen2020powernorm,vaswani2017attention}.

Cross-Iteration BatchNorm (CBN)~\citep{yao2021cross}, EvalNorm~\citep{singh2019evalnorm} and others~\citep{ioffe2017batch,yan2020towards,yang2021back} are extensions of BN which address some of these issues. Normalization layers which only operate on single samples (e.g.\ GroupNorm~\citep{wu2018group}, LayerNorm~\citep{ba2016layer}) have also been proposed~\citep{ba2016layer,labatie2021proxy,liu2020evolving,singh2020filter,ulyanov2016instance,wu2018group}. Dynamic Normalization and Relay (DNR)~\citep{cai2021dynamic} proposed recurrent estimation of affine parameters for video models but does not address the discrepancy of batch and global statistics.

\citet{wu2021rethinking} provide the most extensive review of BN issues for vision to date, including train-test discrepancies, the effect of batch sizes and correlated samples as well as solutions such as FrozenBN~\citep{wu2021rethinking} or redefining over which samples to normalize.
Related to our work, they show problems of small batches for estimating batch statistics and discuss cheating behavior in object detection and contrastive learning.
Consequences for video are only hypothesized. Further, \citet{wu2018group} show how BN hides benefits of higher frame rates for clip-based action classification.

Recently, there has been a trend towards CNNs without BN. \citet{brock2021high} discuss disadvantages of BN and propose a class of normalizer-free networks \emph{NFNet}. They, however, require custom optimizers and are sensitive to hyperparameters.  Finally, \emph{ConvNeXt}~\citep{liu2022convnet} is a CNN competitive with Vision Transformers~\citep{dosovitskiy2020image}. It replaces BatchNorm with LayerNorm although this was not the focus of the paper.

\begin{table}[t]
\begin{center}
\caption{
Previous approaches for video-based surgical workflow tasks.
End-to-end approaches on short sequences have been proposed initially. Recent research has favored multi-stage approaches where each model component is trained separately. ({\color{orange}(\cmark)} = contains an end-to-end stage)
}
\label{table:related_work}
\resizebox{\linewidth}{!}{
\begin{tabular}{lcccl}
\hline
Methods & BN & End2End & Train stages & Remarks\\
\hline
\specialcell{AL-DBN\citep{bodenstedt2019active},\\Tool-Ant\citep{rivoir2020rethinking}} & \xmark & \cmark & 1 &
\specialcell{End-to-end CNN-LSTM\\{\color{red}\xmark\hspace{0.2em}} \emph{{\color{red}non-SOTA backbone} (AlexNet)}}\\
\hline
\specialcell{SV-RCNet\citep{dosovitskiy2020image},\\MTRCNet\citep{jin2020multi}} & \color{red}\cmark & \color{orange}(\cmark) & \color{red}2 & \specialcell{CNN-LSTM w/ end-to-end stage\\{\color{red}\xmark\hspace{0.2em}} \emph{many {\color{red}short sequences} per batch}}\\
\hline
\specialcell{TMRNet\citep{jin2021temporal},\\LRTD\citep{shi2020lrtd}} & \color{red}\cmark & \color{orange}(\cmark) & \color{red}2 & \specialcell{SV-RCNet + Attention\\{\color{red}\xmark\hspace{0.2em}} \emph{many {\color{red}short sequences} per batch}}\\
\hline
CataNet\citep{marafioti2021catanet} & \color{red}\cmark & \color{orange}(\cmark) & \color{red}4 & \specialcell{CNN-LSTM w/ end-to-end stage\\{\color{red}\xmark\hspace{0.2em}} \emph{requires {\color{red}4 training stages}}}\\
\hline
\specialcell{TransSVNet\citep{gao2021trans},\\TeCNO\citep{czempiel2020tecno}\\and others
} & \color{red}\cmark & \color{red}\xmark & \color{red}2-3 & \specialcell{CNN+\{RNN,TCN,Transformer\}\\{\color{red}\xmark\hspace{0.2em} \emph{separate training stages}}\\\emph{for each model component}}\\
\hline
Proposed & \xmark & \cmark & 1 & \specialcell{End-to-end CNN-LSTM\\\emph{SOTA using BN-free backbones}}\\
\hline
\end{tabular}
}
\end{center}
\end{table}

\subsection{Online Surgical Workflow Analysis}

\begin{figure}[t]
    \centering
    \includegraphics[trim=0cm 9.8cm 12.5cm .1cm,clip,width=\linewidth]{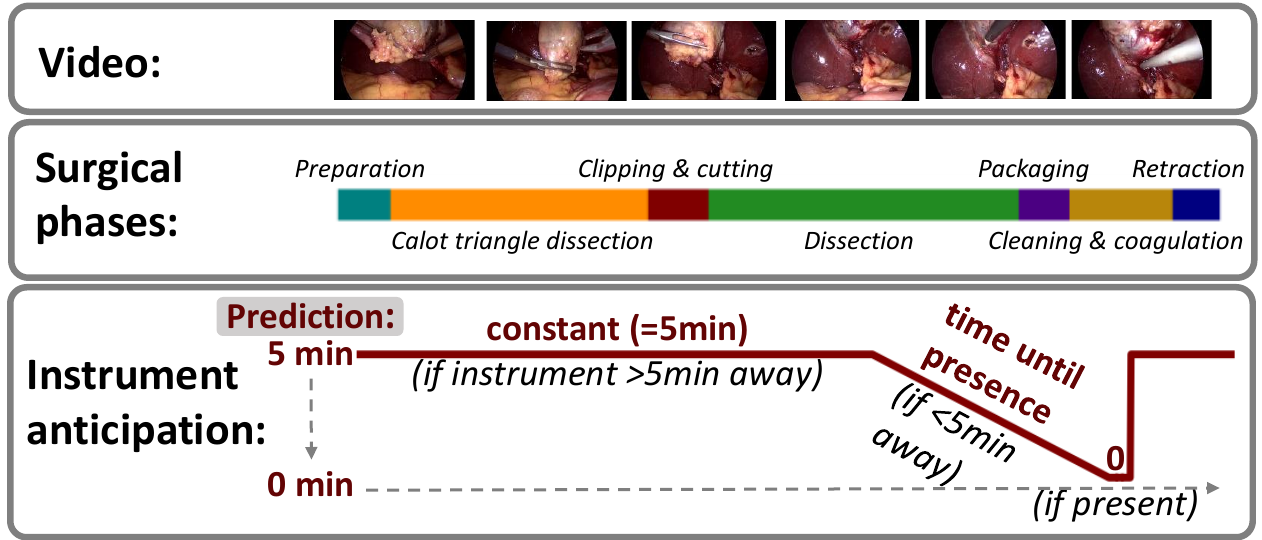}
    \caption{Two surgical workflow tasks on the Cholec80 dataset~\citep{twinanda2016endonet}: {\color{\colorrevtwo}\emph{phase recognition}, a special case of temporal action segmentation, and \emph{instrument anticipation}, defined as predicting the time until occurrence of an instrument within a specified horizon.}}
    \label{fig:task}
\end{figure}

Surgical Workflow Analysis (SWA)~\citep{maier2017surgical} is a broad set of tasks which aim at understanding events and their interrelations in recorded or live surgical videos {\color{\colorrevtwo}(Fig.~\ref{fig:task})}. 
SWA covers a wide range of common computer vision problems including temporal action segmentation (e.g.\ surgical phase recognition~\citep{twinanda2016endonet}), dense frame-wise regression (e.g.\ anticipating instrument usage~\citep{rivoir2020rethinking} or procedure duration~\citep{twinanda2018rsdnet}) or video classification (e.g.\ early surgery type recognition~\citep{kannan2019future}). 
SWA tasks are inherently video-based as they require understanding the sequence of events which have occurred until the current time point and often require memorizing long-range dependencies.
The overall goal is to provide live, context-aware assistance during surgery~\citep{maier2022surgical}. Thus, tasks are mostly defined as online recognition problems~\citep{czempiel2020tecno,kannan2019future,rivoir2020rethinking,twinanda2018rsdnet}.

Most approaches to surgical workflow tasks combine a 2D CNN for frame-wise feature extraction with a temporal model (e.g.\ LSTM, TCN or Attention) for aggregation over time. 
In the following, we discuss different learning strategies which have emerged for surgical workflow tasks and how they were possibly influenced by BN-related problems. Table \ref{table:related_work} provides an overview of our findings.

\textbf{E2E without BN:}
Early methods completely avoid BN-issues by using backbones without BatchNorm. 
End-to-end CNN-LSTM models with AlexNet~\citep{krizhevsky2012imagenet} or VGG~\citep{simonyan2014very} backbones
have been proposed for phase recognition~\citep{bodenstedt2019active,yengera2018less}, anticipation~\citep{rivoir2020rethinking}, duration prediction~\citep{rivoir2019unsupervised,yengera2018less}, 
surgery type prediction~\citep{kannan2019future} and occlusion detection~\citep{bano2020fetnet}. All methods train long sequences of several minutes. However, results are not competitive due to the use of outdated backbones.

\textbf{E2E with BN:}
SV-RCNet~\citep{jin2017sv} and MTRCNet~\citep{jin2020multi} are 
two examples of 
end-to-end ResNet-LSTM models for phase recognition. However, to minimize the discrepancy between batch and global statistics, training sequences are only 10 frames long (2 and 10 seconds, respectively) and processed in large batches of 100 sequences.
TMRNet~\citep{jin2021temporal} and LRTD~\citep{shi2020lrtd} extend SV-RCNet with attention blocks to learn long-range dependencies. Still however, sequences of only 10 frames are trained end to end. CataNet~\citep{marafioti2021catanet} proposes a complex, 4-stage learning process for ResNet-LSTMs to predict surgery duration.
In one of the stages, end-to-end learning on long sequences is made possible by using FrozenBN with global statistics from a previous stage.

\textbf{Multi-stage:} Recently, methods have shifted towards more complex temporal models on top of pretrained visual features. BN-related issues are circumvented by first training a CNN on randomly sampled image batches, followed by the temporal model trained on frozen image features.
Methods in this style have been proposed for phase recognition~\citep{czempiel2020tecno,czempiel2021opera,gao2021trans,yi2022not,zisimopoulos2018deepphase}, duration prediction~\citep{aksamentov2017deep,bodenstedt2019prediction,twinanda2018rsdnet}, tracking~\citep{nwoye2019weakly} or anticipation~\citep{yuan2021surgical}. Most notably, TeCNO~\citep{czempiel2020tecno}, an MS-TCN~\citep{farha2019ms} trained on ResNet features, is the popular approach for 2-stage learning and Trans-SVNet~\citep{gao2021trans}, a 3-stage method which trains a Transformer on TeCNO features, is often considered state of the art among methods with CNN backbones. More recently, few multi-stage methods have achieved strong performance by using Transformer-based backbones~\citep{chen2022spatio,he2022empirical,liu2023lovit,liu2023skit}.

\textbf{Awareness for BN issues:} Although BatchNorm has possibly influenced the move towards complex, multi-stage or short-range models, its issues have rarely been discussed. In their end-to-end model, \citet{yengera2018less} justify the use of the BN-free AlexNet with efficiency reasons. \citet{kannan2019future} simply state that their end-to-end model with VGG outperformed ResNet, but not whether this may be due the absence to BatchNorm. The authors of SV-RCNet~\citep{jin2017sv} do not explain their choice of short sequences 
despite large batch sizes. \citet{nwoye2019weakly} justify their 2-stage approach through ``fair comparison''. CataNet~\citep{marafioti2021catanet} uses FrozenBN in an end-to-end learning stage~\footnote{\url{https://github.com/aimi-lab/catanet/blob/main/train\_rnn.py\#L179}}, but this is not mentioned in the paper. \citet{rivoir2020rethinking} briefly mention a cheating phenomenon to justify the choice for an AlexNet in instrument anticipation.

\textbf{3D backbones:}
It is noticable that surgical workflow methods almost unanimously use 2D backbones as feature extractors. The specialized surgical domain requires backbones to be finetuned and several studies suggest that the small-scale public datasets available in this domain are not sufficient to train large 3D CNNs~\citep{czempiel2022surgical,zhang2022large}. Only few works achieve good results on larger private datasets~\citep{czempiel2022surgical,zhang2021swnet}. More recently, \citet{he2022empirical} confirmed poor performance of 3D CNNs but showed promising results using the spatio-temporal Transformer architecture \emph{Video Swin}~\citep{liu2022video} on short 1-sec.\ sequences.

\subsection{Action Segmentation \& Anticipation}

In the natural-video domain, action segmentation and anticipation can be viewed as workflow analysis tasks as they aim at understanding the composition of complex activities.

\textbf{Training strategies:} Methods avoid BatchNorm during temporal learning by almost exclusively relying on pretrained visual features~\citep{carreira2017quo} in both action segmentation~\citep{farha2019ms,huang2020improving,ishikawa2021alleviating,wang2020boundary,chinayi_ASformer} and anticipation~\citep{abu2020long,furnari2020rolling,gao2017red,ke2019time,sener2020temporal}.
Whether this development can be attributed to computational costs, the availability of well-pretrained feature extractors or BatchNorm's properties is not yet clear.

\citet{sener2020temporal} indicate limitations of pretrained visual features for action anticipation by showing a large performance gap compared to using ground-truth observed actions as input.
The only end-to-end anticipation model for natural video of which we are aware (AVT,~\citep{girdhar2021anticipative}) does not use BatchNorm as it is entirely Transformer-based. Although not mentioned as motivation for this model choice, BN would likely have caused both train-test discrepancy and ``cheating'' since single-sequence batches are used.
Recently, \citet{liu2022empirical} show the effectiveness of end-to-end strategies for the related temporal action detection task. They use FrozenBN but do not discuss BatchNorm issues.

\textbf{Online recognition:}
Online action understanding has mostly focused on temporal action detection~\citep{eun2020learning,lea2017temporal,xu2019temporal,zhao2020privileged} or anticipation. Apart from few exceptions~\citep{ghoddoosian2022weakly}, action segmentation is usually done offline~\citep{farha2019ms,huang2020improving,ishikawa2021alleviating,wang2020boundary,chinayi_ASformer}.

\section{Making End-to-End Video Learning Work}

We analyze limitations and impact of BatchNorm for end-to-end learning on two surgical tasks (Fig.~\ref{fig:task}). In \emph{surgical phase recognition}, an online temporal segmentation task, we show how BatchNorm can affect training strategies and performance. Especially, we show that supposedly outdated CNN-LSTMs can be very effective when BN's pitfalls are taken into consideration. Then, we demonstrate BatchNorm's ability to access future frames in online tasks through a ``cheating'' phenomenon in \emph{surgical instrument anticipation}.

\subsection{Proposed End-to-End Learning Strategies}
Our premise is that end-to-end learning is preferable over multi-stage approaches and thus we propose a simple, intuitive strategy for training CNN-LSTM models on online surgical workflow tasks. We use the LSTM to demonstrate that even simple temporal aggregation can be effective when visual features are improved through end-to-end learning {\color{\colorrevone}and temporal context is maximized}. We also argue that recurrent, state-based models, where a memory state propagates information through time, are often well suited for online recognition tasks~\citep{furnari2020rolling,gao2017red,ghoddoosian2022weakly}. Especially for end-to-end learning on partial video sequences, the LSTM's hidden state can be utilized to increase temporal context.
Specifically, we propose:

\textbf{End-to-end training:} We optimize the feature backbone and temporal model end to end in a single training stage.

\textbf{Single-sequence batch:} {\color{\colorrevtwo}In end-to-end settings, batches are limited to partial clips containing only few minutes of total video due to memory constraints. We maximize temporal context by using a single sequence/clip of 64-256 frames per batch.}

\textbf{Carry-hidden training (CHT) \emph{(optional)}:} We select non-overlapping batches in temporal order. This way, the LSTM's detached hidden state can be carried across batches to further increase temporal context during training.

\textbf{Partial freezing \emph{(optional)}:} We freeze bottom layers of the backbone to increase the length of training sequences. Although this is not fully end to end anymore, the main benefits are still given: There is still only a single training stage and image features are optimized in a temporal context.

{\color{\colorrevone}End-to-end learning~\citep{jin2016endorcn,jin2017sv}, single-sequence batches~\citep{kannan2019future} and CHT~\citep{nwoye2019weakly} have been used individually in BN-based approaches. However, only methods with BN-free backbones such as AlexNet have been able to employ these~\citep{bodenstedt2019active} or similar~\citep{yengera2018less} strategies in combination. This is because end-to-end training is not compatible with single-sequence batches if the backbone contains BatchNorm, and by extension with CHT as it requires single-sequence batches. The following paragraph elaborates the main problems.}

\subsection{BatchNorm's Problems with Sequential Data}

\textbf{1) Small-batch effect} (Sec.~\ref{sec:phase}, Fig.~\ref{fig:e2e_2stage},~\ref{fig:train_strategies}):
It is well-known that BatchNorm performs poorly with small batches~\citep{wu2018group,wu2021rethinking}. Batch statistics, used to normalize features during training, only poorly estimate global statistics used during inference and thus induce a discrepancy between train and test behavior. However, when batches contain sequential video frames, even large batches lead to similar issues as the highly correlated samples also do not approximate global statistics well. Similar observations have been made in different contexts~\citep{ba2016layer,wu2018group}.

\textbf{2) Cheating/leakage} (Sec.\ref{sec:anticipation}, Fig.~\ref{fig:toy},~\ref{fig:cheating_anticipation}):
Previous work has 
discussed 
BatchNorm's ability to leak information from other samples in the batch and its impact on performance in object detection or contrastive learning~\citep{wu2021rethinking}. In online video tasks, a similar phenomenon can be fatal. BatchNorm can leak information from future frames, which allows models to ``cheat'' certain objectives and prevents learning useful features.
This ``cheating'' phenomenon is most obvious in anticipation tasks.
In \emph{instrument anticipation}, the occurrence of an instrument
influences the batch statistics of channels reacting to its appearance.
During training, BN models may 
pick up on these fluctuations in batch statistics even at \emph{earlier time points} within the same 
batch and use them 
to anticipate the instrument's later occurrence. Thus, the model only has to learn to \emph{recognize} the instrument to solve the \emph{anticipation} objective during training, but will fail during inference when batch statistics are not available.

\subsection{Alternative Backbones without BatchNorm}
We enable end-to-end learning with single-sequence batches by replacing ResNet50 with existing BN-free CNNs. We use a ResNet50 with \textbf{GroupNorm (GN)}~\citep{wu2018group} for fair comparison as well as the more recent \textbf{ConvNeXt-T}~\citep{liu2022convnet}, which uses ViT-style LayerNorm~\citep{yao2021leveraging} and is of comparable size to ResNet50.
We also investigate \textbf{FrozenBN}~\citep{liu2022empirical,wu2021rethinking}, where batch statistics are replaced by global statistics from the pretraining stage.
All models are pretrained on ImageNet-1k~\citep{krizhevsky2012imagenet}.

{\color{\colorrevtwo}\subsection{Limitations of Previous Strategies}
Existing BatchNorm-based approaches avoid issues with sequential data either through multi-stage training or multi-sequence batches. Both solutions, however, suffer from potential limitations which we attempt to explain in the following. We support these claims through our experiments.}

\textbf{Multi-stage:} {\color{\colorrevtwo}Multi-stage training is the most popular approach and effectively avoids BN's issues by training backbones without temporal context.} Yet, this has several disadvantages. Firstly, it increases the number of hyperparameters since learning rate, number of epochs etc. have to be tuned for each training stage. Secondly, when backbone and temporal model are trained on the same data, overfitting on the image task can prevent the temporal model from learning useful features. For example, we found early stopping of the ResNet pretraining crucial to reproduce TeCNO's~\citep{czempiel2020tecno} results.
Lastly and most importantly, pretraining on single frames changes the dynamics of the task and cannot guarantee features that are suitable for the target task. A certain visual feature might only be relevant to the task in combination with a previous event. If the backbone does not learn to capture this feature during pretraining, this information is lost and cannot be recovered in the subsequent temporal-modeling stage.

\textbf{End-to-end with multi-sequence batches:} {\color{\colorrevtwo}Sampling multiple sequences per batch (e.g.\ SV-RCNet~\citep{jin2017sv}) reduces the discrepancy between batch and global statistics.} However, this comes at the cost of shorter training sequences. It also restricts possible training strategies like carrying hidden states of LSTMs across batches and can thus impair performance.






\section{Experiments} \label{sec:experiments}

\begin{figure*}
\centering
\begin{minipage}{.24\textwidth}
    \centering
    \includegraphics[trim=.3cm .2cm .25cm .2cm,clip,width=.88\linewidth]{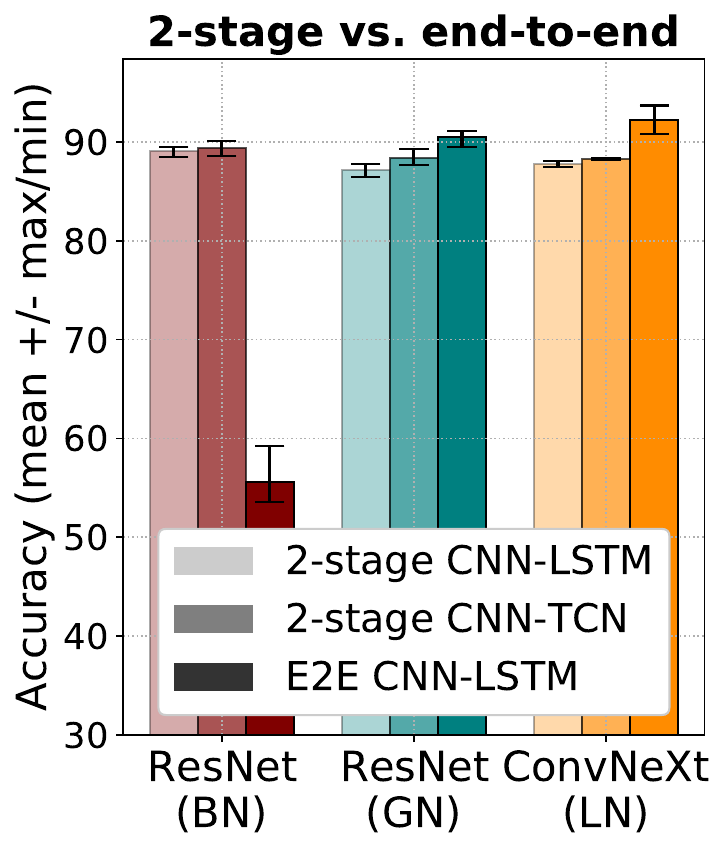}
    \caption{\emph{Phase Recognition (Cholec80):} BN-free, end-to-end CNN-LSTMs outperform 2-stage LSTMs and TCNs. The widely-used ResNet50 with BN fails. Split: 40/8/32 from TeCNO~\citep{czempiel2020tecno}.}
    \label{fig:e2e_2stage}
\end{minipage}
\hfill
\begin{minipage}{.73\textwidth}
        \centering
        \begin{subfigure}[b]{0.27\linewidth}
            \centering
            \includegraphics[trim=.2cm .2cm .2cm .2cm,clip,height=3.8cm]{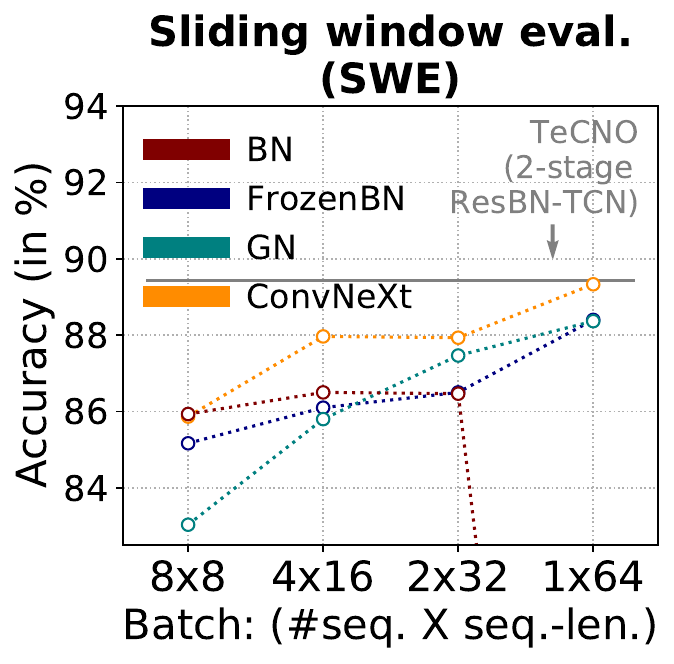}
            \caption{\emph{Length vs diversity}}
            \label{fig:seq_len}
        \end{subfigure}    
        \hfill
        \begin{subfigure}[b]{0.43\linewidth}
            \centering
            \includegraphics[trim=.2cm .2cm .2cm .2cm,clip,height=3.8cm]{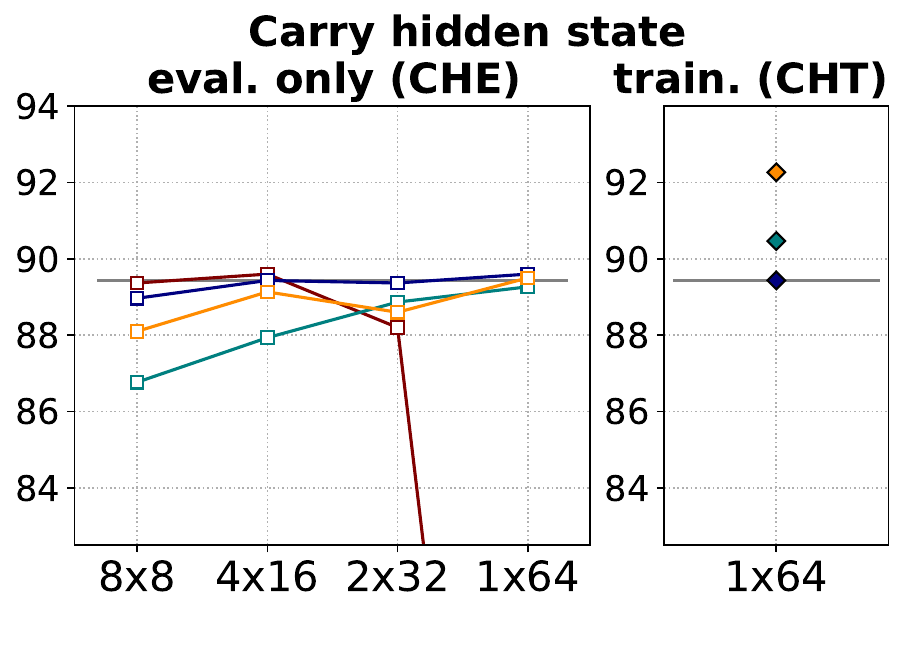}
            \caption{\emph{CHT strategy improves BN-free models}}
            \label{fig:carry_hidden}
        \end{subfigure}
        \hfill
        \begin{subfigure}[b]{0.22\linewidth}
            \centering
            \includegraphics[trim=.2cm .2cm .2cm .2cm,clip,height=3.8cm]{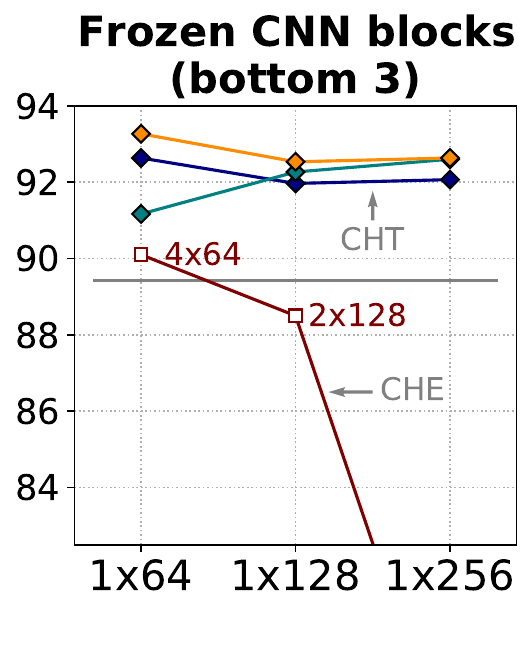}
            \caption{\emph{Partial freezing}}
            \label{fig:frozen}
        \end{subfigure}
        \caption{\emph{Phase Recognition (Cholec80) with end-to-end CNN-LSTMs:} (a) While longer sequences improve BN-free models, BN in ResNet50 backbones requires careful selection of the number of sequences per batch before performance drops. (b) 
        Carrying hidden state through entire videos at test time improves models. By also doing this during training, train-test discrepancy is removed and models improve further. In BN models, 
        implementing this is not straightforward. 
        (c) BN models improve by freezing backbone blocks and thus increasing training sequences while maintaining batch diversity.  
        Yet, freezing also improves BN-free models. Explicit scores with standard deviations {\color{\colorrevone}and additional metrics can be found in the appendix (Tables~\ref{table:acc}~\&~\ref{table:f1})}.}
        \label{fig:train_strategies}
\end{minipage}
\end{figure*}

We provide a comprehensive and detailed analysis of how BatchNorm affects end-to-end surgical workflow analysis. We show the advantage of end-to-end over 2-stage learning \emph{(Hypothesis 1)}, longer training sequences \emph{(H.2)} and carrying hidden states across batches in online tasks \emph{(H.3)} and how this can fail using BN-based backbones. We confirm our findings on similar, natural-video datasets \emph{(H.4)} and other BN variants \emph{(H.5)}. Then, we discuss how freezing layers can alleviate BN-issues \emph{(H.6)} and position our proposed models within the state of the art for surgical phase recognition \emph{(H.7)}.
We show BN's potential to cheat \emph{(H.8)}, provide evidence in instrument anticipation \emph{(H.9)} and compare models to previous anticipation approaches \emph{(H.10)}.

\subsection{Data, Tasks \& Implementation Details}

{\color{\colorrevtwo}\textbf{Cholec80:}} For our main analysis, we train phase recognition and anticipation models on the widely-studied \emph{Cholec80 dataset}~\citep{twinanda2016endonet}, which consists of 80 recorded gallbladder removals (cholecystectomies). Videos range from 12 to 100\thinspace min (mean ca.\ 38\thinspace min), processed at 1fps with frame-wise labels for 7 instruments and 7 surgical phases.

\emph{Surgical phase recognition} is an dense online temporal action segmentation task~\citep{twinanda2016endonet}. We use the mean video-wise accuracy for evaluation (F1 in Appendix Table~\ref{table:f1}).

\emph{Instrument anticipation} is an online frame-wise regression task. At each time point, the objective is to predict the remaining time until occurrence of an instrument within a horizon of 5 minutes. Outside the horizon, a constant should be predicted (see Fig.~\ref{fig:task}). We follow the recently proposed task formulation~\citep{rivoir2020rethinking}. We use the weighted MAE (wMAE) as evaluation metric, where errors inside and outside the horizon are weighted equally to counteract imbalances.

For both tasks, we use 40 videos for training, 8 for validation and 32 for testing, following TeCNO~\citep{czempiel2020tecno}. Epochs with the best validation score (\emph{acc./wMAE}) are selected for testing. For SOTA comparison in phase recognition, we retrain models on a 32/8/40 split. This allows us to stay comparable to the widely used 40/40 split~\citep{twinanda2016endonet,gao2021trans} while still having a validation set for model selection. All runs are repeated 3 times. More training and evaluation details can be found in \ref{sec:hyp} \& \ref{sec:metrics}.

{\color{\colorrevtwo}\textbf{GTEA \& 50Salads:} To reproduce our claims on natural videos, we retrain phase recognition models for online action segmentation on \emph{GTEA}~\citep{fathi2011learning} and \emph{50Salads}~\citep{stein2013combining}.}

{\color{\colorrevtwo}\textbf{AutoLaparo \& CATARACTS:} For state-of-the-art comparisons, we retrain phase recognition models on two additional surgical datasets. AutoLaparo~\citep{wang2022autolaparo} contains 21 long videos (10/4/7 split) ranging from 27 to 112\thinspace min (mean ca.\ 66\thinspace min) with 7 coarse phases annotated. CATARACTS~\citep{zisimopoulos2018deepphase} contains 50 short videos ranging from 6 to 40\thinspace min (mean ca.\ 11\thinspace min) with 19 fine-grained steps annotated. To be able to compare to the 25/25 split used in previous work~\citep{chen2022spatio}, we use a 20/5/25 split. On both datasets, we report the mean video-based accuracy as well as macro-precision, -recall and -jaccard, which are computed over all test frames. See \ref{sec:hyp} \& \ref{sec:metrics} for details.}

\subsection{Surgical Phase Recognition} \label{sec:phase}

{\bf Hypothesis 1: End-to-end outperforms 2-stage training, but not with BatchNorm.}
We argue that end-to-end learning is preferable over 2-stage approaches but fails when the backbone contains BatchNorm. To demonstrate this, we train CNN-LSTM models with three different backbones (Resnet50-BN, ResNet50-GN and ConvNeXt-T), single-sequence batches of 64 frames ($1\times64$) and the proposed carry-hidden training (CHT). For comparison, we train each backbone in a 2-stage process with an LSTM or causal MS-TCN, following TeCNO's training procedure~\citep{czempiel2020tecno}.

Fig.~\ref{fig:e2e_2stage} shows that end-to-end models outperform 2-stage approaches when using a backbone without BN. Only the 
typically used
BN-based ResNet50
fails completely.
While BatchNorm is known to fail with small batch sizes~\citep{wu2018group,wu2021rethinking}, even large batch sizes lead to similar issues here due to the strong correlation of subsequent frames. In other words, a batch with a single 64-frame sequence behaves similar to a batch size of 1. The next hypothesis investigates this in more detail. 

{\bf Hypothesis 2: BatchNorm causes trade-off between sequence length and batch diversity.}
We investigate the effect of different batch configurations and follow SV-RCNet's~\citep{jin2017sv} strategy for end-to-end CNN-LSTMs: Multiple shorter sequences are sampled per batch, hidden states are reset after each training sequence and the LSTM uses a sliding window for evaluation \emph{(SWE)} to ensure consistent sequence lengths at train and test time. We test this strategy using 1-8 sequences in batches of 64 frames in total.

In Fig.~\ref{fig:seq_len}, we observe that the performance of all BN-free models improves with fewer but longer sequences per batch since it increases the temporal context.
In BN models, however, fewer sequences poorly approximate global statistics, {\color{\colorrevtwo}leading to a discrepancy in feature normalization at train and test time.} Thus, we observe drops in performance despite the constant total batch size, similar to the frame-rate trade-off in video-clip classification~\citep{wu2018group}.
The BN backbone outperforms its FrozenBN and GroupNorm counterparts for multiple short sequences but stagnates and then collapses with fewer, longer ones. Due to this trade-off between sequence length and batch diversity, BN models eventually cannot reach the performance of the fully-sequential, BN-free models.

\begin{figure}[t]
    \centering
    \includegraphics[trim=.2cm .2cm .2cm .2cm,clip,width=.85\linewidth]{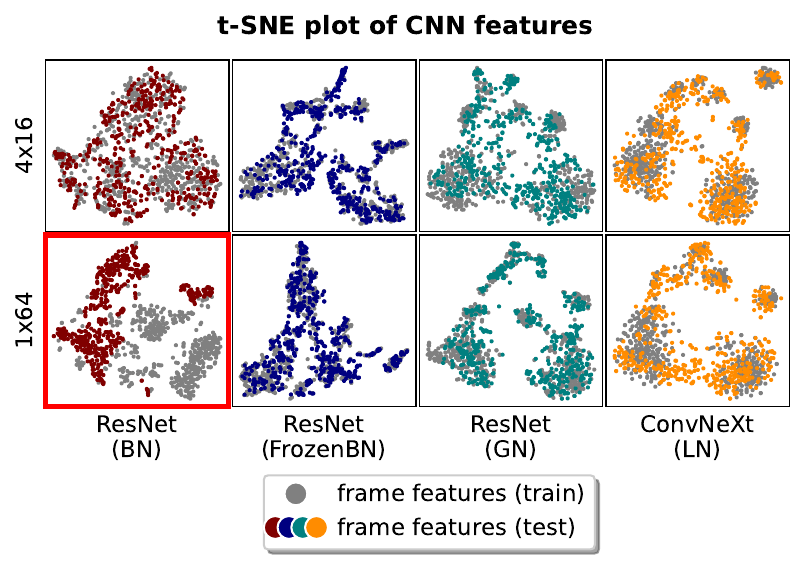}
    \caption{{\color{\colorrevtwo}\emph{Phase recognition (Cholec80):} With multi-sequence batches (4x16), train and test features cover similar distributions. In BN-based backbones, single-sequence batches (1x64) cause a shift in train and test features due to the discrepancy between batch and global statistics.}}
    \label{fig:tsne}
\end{figure}

{\color{\colorrevtwo}To show BatchNorm's train-test discrepancy, we visualize CNN features in Fig.~\ref{fig:tsne} with t-SNE plots (perplexity: $30$). When training with multiple short sequences per batch (4x16), train and test features cover very similar distributions for all backbones. With single-sequence batches (1x64), however, we observe a shift between train and test features for BN-based backbones, while this issue does not occur with BN-free backbones.}

{\bf Hypothesis 3: BatchNorm restricts LSTM usage for online tasks.}
Note that by adopting SV-RCNet's hidden-state handling, none of the methods outperform the 2-stage SOTA~\citep{czempiel2020tecno}. This is likely due to the short temporal context seen by the LSTM. Simply carrying the hidden state across the entire video during inference \emph{(CHE)} already provides a clear performance boost without changing the training strategy (Fig.~\ref{fig:train_strategies}, \emph{SWE vs. CHE}).
Yet, this induces another train-test discrepancy since the LSTM accumulates hidden states over much longer time periods at test time compared to training time, which could cause unpredictable behavior.

In online tasks, however, the discrepancy can naturally be resolved by exploiting the fact that hidden states only flow in a single direction. Specifically, we can implement the proposed \emph{carry-hidden training (CHT)} by selecting batches in temporal order and carrying the detached hidden states across batches during training as well.

With this approach, {\color{\colorrevone}temporal context is increased during training as well as inference. Now,} GN and ConvNeXt backbones outperform 2-stage and other end-to-end strategies (Fig.~\ref{fig:carry_hidden}). In BN models, however, this learning strategy is not implementable in a straightforward way since it requires selecting batches in temporal order, which violates the i.i.d. requirement. Thus, the train-test discrepancy in LSTMs remains.

Interestingly, the \emph{CHT} strategy is not effective with FrozenBN. We found models often to collapse and finally only achieve subpar performance by lowering the initial learning rate. The lack of proper normalization with FrozenBN might cause more instability than in GN or ConvNeXt models and we suspect that the non-i.i.d. selection of batches during \emph{CHT} can encourage this. We present more evidence in our experiments on other datasets (\emph{Hypothesis 4}, Fig.~\ref{fig:other}) and a more detailed exploration of learning rates in the appendix (Table~\ref{table:lr}).


\begin{figure}[t]
    \centering
    \includegraphics[trim=.2cm .2cm .2cm .2cm,clip,width=.89\linewidth]{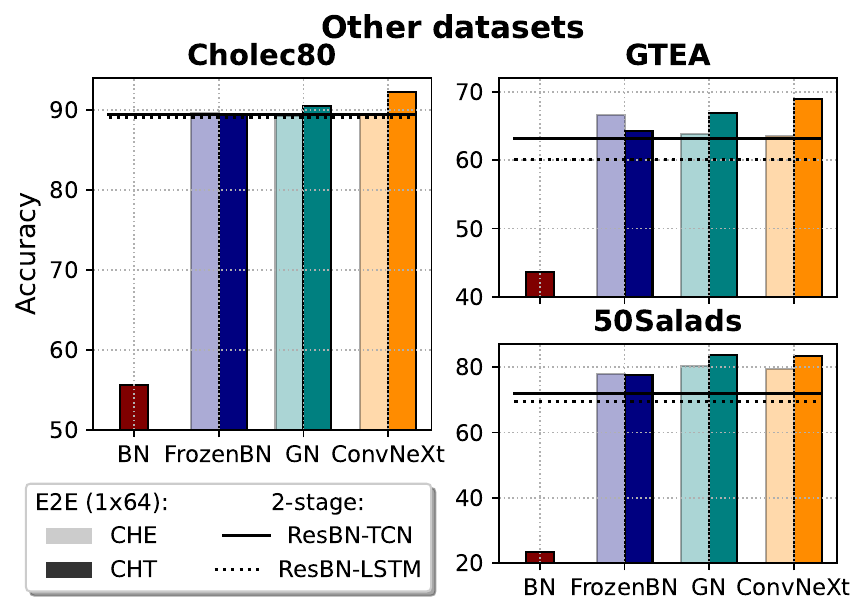}
    \caption{\emph{Online action segmentation with end-to-end CNN-LSTMs:} Results on surgical data (left) can be reproduced on natural-video datasets (right).}
    \label{fig:other}
\end{figure}

{\bf Hypothesis 4: BatchNorm induces similar issues on natural-video datasets.}
Fig.~\ref{fig:other} confirms our findings on small-scale datasets from the natural-video domain (\emph{GTEA} and \emph{50Salads}) by retraining the same models used for phase recognition on online action segmentation. We find that our main claims can be reproduced, namely that (1) BN-based end-to-end models fail with single-sequence batches, (2) end-to-end approaches without BN outperform 2-stage models and (3) the CHT strategy is effective using GN or ConvNeXt but not with FrozenBN. {\color{\colorrevone}Our proposed strategies aim at increasing temporal context, and thus seem mostly applicable to tasks where this is relevant (e.g. long videos and workflow-based tasks such as cooking or other complex activities). Its impact on other video tasks remains open.}

{\bf Hypothesis 5: Other BatchNorm variants are not suitable.}
In Table~\ref{table:other_norms}, {\color{\colorrevone} we perform a more extensive comparison of batch-independent and batch-dependent normalization layers. We find that other batch-independent methods (InstanceNorm, LayerNorm) benefit from our context-enlarging strategies. However, even BatchNorm variants which attempt to alleviate issues with temporal dependence or small batches exhibit similar failure cases as BatchNorm. Temporal dependence can be avoided entirely by normalizing each time step individually. Yet, with single-sequence batches this still fails due to normalization over single frames.} DNR~\citep{cai2021dynamic} proposes temporally dependent estimation of BN's affine parameters to adjust to properties of batches in video settings. Yet, batch stats are still computed like in standard BN, so the train/test discrepancy with single-sequence batches remains and model collapse with 1x64 batches can clearly be observed. CBN~\citep{yao2021cross} reduces batch dependence by accumulating stats from multiple batches. Yet, for our proposed CHT strategy, CBN is not suitable due to strong inter-batch correlations of subsequent batches. Further, we found the transition to accumulated stats after its burn-in phase to cause instability even in CHE settings, despite finetuning. Expectedly, this issue was stronger in the single-sequence case (1x64, CHE).

\begin{table}[t]
\begin{center}
\caption{\emph{Phase recognition (Cholec80):} {\color{\colorrevone}Comparison of batch-independent and -dependent normalization techniques.}}
\label{table:other_norms}
\resizebox{\linewidth}{!}{
\begin{tabular}{l|c|c|c|}
\hline
\multicolumn{4}{c}{\emph Accuracy\ (mean$\pm$std.\ over runs) on 40/8/32 split from TeCNO~\citep{czempiel2020tecno}.}\\
\hline
\emph{End-to-end ResNet-LSTM w/ ...} & $4\times16$, CHE & $1\times64$, CHE & $1\times64$, CHT\\
\hline
Batch-\textbf{independent} normalization & \multicolumn{3}{c|}{} \\
\hline
\color{\colorrevone}InstanceNorm~\citep{ulyanov2016instance} & $87.3 (\pm2.5)$ & $88.5 (\pm0.5)$ & $\bf90.2 (\pm0.2)$ \\
\color{\colorrevone}LayerNorm~\citep{ba2016layer} & $87.6 (\pm1.6)$ & $89.1 (\pm1.3)$ & $\bf90.4 (\pm2.8)$ \\
GroupNorm~\citep{wu2018group} & $87.9 (\pm0.8)$ & $89.3 (\pm0.4)$ & $\bf90.5 (\pm0.9)$ \\
\hline
Batch-\textbf{dependent} normalization & \multicolumn{3}{c|}{} \\
\hline
BatchNorm w/ norm.\ over dim.\ B,T,H,W (standard)  & $89.6 (\pm1.0)$ & {\color{red}$59.7 (\pm3.8)$} & {\color{red}$55.6 (\pm3.1)$} \\
\color{\colorrevone}BatchNorm w/o norm.\ over temp.\ dim.\ T (only B,H,W) & $88.4 (\pm0.5)$ & {\color{red}$62.7 (\pm7.1)$} & {\color{red}$44.6 (\pm0.7)$} \\
DNR~\citep{cai2021dynamic} & $88.9 (\pm0.7)$ & {\color{red}$68.0 (\pm6.5)$} & {\color{red}$61.6 (\pm2.7)$} \\
CBN~\citep{yao2021cross} w/ window size 1 \emph{(equiv.\ to BN)} & $89.5 (\pm0.7)$ &  &   \\
CBN w/ burn-in 3 epochs, window size 8 \emph{(original)} & {\color{red}$49.1 (\pm2.4)$} & & \\
CBN w/ burn-in 10 epochs, window size 4 & $87.4 (\pm0.8)$ & $79.4 (\pm1.3)$ & {\color{red}$51.5 (\pm19.8)$} \\
\hline
\end{tabular}
}
\end{center}
\end{table}

{\bf Hypothesis 6: Freezing backbone layers can improve BN models, but also their BN-free counterparts.}
As argued before, the trade-off between sequence length and batch diversity is a disadvantage of BN-based models. By freezing parts of the CNN backbones, sequence lengths can be increased while maintaining the same number of sequences. Hence, we freeze the bottom three blocks of all backbones and retrain each model with sequence lengths of 64, 128 and 256. For BN-free models, we choose the CHT strategy but resort to CHE for the BN model (Fig.~\ref{fig:frozen}).

Through freezing, 4x64-BN finally outperforms the 2-stage SOTA but collapses again with longer sequences while GN models improve. Even FrozenBN matches other BN-free models. Consistent with prior beliefs~\citep{he2016deep}, lack of normalization might be less problematic in shallow models.

\begin{table*}[t]
\begin{center}
\caption{State of the art in surgical phase recognition on Cholec80~\citep{twinanda2016endonet} with a 40/40 split. Our models are trained on a 32/8/40 split to allow for model selection. We find that simple, end-to-end CNN-LSTMs match or outperform multi-stage methods with attention-based temporal models~\citep{gao2021trans,jin2021temporal,chen2022spatio}, complex augmentation pipelines~\citep{yi2022not} or Transformer backbones~\citep{chen2022spatio,he2022empirical}.}
\label{table:sota}
\resizebox{.9999\linewidth}{!}{
\begin{tabular}{lllllllll}
\hline
\multicolumn{9}{c}{\textbf{Cholec80}} \\
\hline
Method & Backbone & Temporal & \#Training & relaxed Accuracy & relaxed Jaccard & Accuracy & \color{\colorrevone}Jaccard & Balanced Accuracy\\
& & Model & Stages & (std. over videos) & (std. over phases) & (std. over runs) & (std. over runs) & (std. over runs)\\
\hline
SV-RCNet~\citep{jin2017sv} \emph{(E2E, 100x10, SWE)} & ResNet-BN & LSTM & 2 & $85.3 (\pm7.3)$ & & $85.0$ & $65.5$ \\
TeCNO~\citep{czempiel2020tecno} & ResNet-BN & TCN & 2 & $88.6 (\pm7.8)$ & $75.1 (\pm6.9)$ & $88.1 (\pm0.3)^\dagger$ & $69.4 (\pm0.5)^\dagger$ & $85.4 (\pm1.0)^\dagger$ \\
TMRNet~\citep{jin2021temporal} & ResNet-BN & LSTM, Attn. & 2 & $89.2 (\pm9.4)$ & $78.9 (\pm5.8)$ \\
Trans-SVNet~\citep{gao2021trans} & ResNet-BN & TCN, Transf. & 3 & $90.1 (\pm7.1)$ & $79.3 (\pm6.6)$ & $89.0 (\pm1.1)^\dagger$ & $71.5(\pm2.4)^\dagger$ & $85.4(\pm1.4)^\dagger$ \\
Not-E2E~\citep{yi2022not} & ResNet-BN & TCN, GRU & 3 & $92.0 (\pm5.3)$ & $77.1 (\pm11.5)$ \\
\hline
TMRNet~\citep{jin2021temporal} & ResNeST & LSTM, Attn. & 2 & $90.1 (\pm7.6)$ & $79.1 (\pm5.7)$ \\
PATG~\citep{kadkhodamohammadi2022patg} & SE-ResNet & GNN & 2 & & & $91.4$ \\
\hline
\citet{zhang2022large} & C3D & LSTM & 2 & $85.9 (\pm7.9)$ \\
\citet{he2022empirical} & I3D & GRU & 2 & & & $88.3 (\pm1.0)$ \\
\hline
TeSTra~\citep{zhao2022real} & Transf. & Transf. & 2 & & & $90.1$ & $71.6$ \\
\citet{he2022empirical} & Video-Transf. & GRU & 2 & & & $90.9 (\pm0.01)$ \\
Dual Pyramid~\citep{chen2022spatio} & Transf. & Transf. & 2 & & & $91.4$ & $75.4$ \\
\hline
\hline
\emph{\textbf{Ours w/ BN}} \\
\hline
E2E (4x16, SWE) \emph{(comparable to SV-RCNet)} & ResNet-BN & LSTM & 1 & $87.5 (\pm7.7)$ & $73.1 (\pm8.5)$ & $86.6 (\pm1.1)$ & $66.3 (\pm0.8)$ & $82.9 (\pm0.6)$ \\
E2E (4x16, CHE) &  &  & 1 & $90.3 (\pm7.1)$ & $77.5 (\pm8.5)$ & $89.3 (\pm0.7)$ & $70.4 (\pm1.0)$ & $84.9 (\pm0.8)$\\
Partial freezing (4x64, CHE) &  &  & 1 & $91.3 (\pm6.9)$ & $\bf{79.8}(\pm9.2)$ & $90.1 (\pm0.5)$ & $72.8 (\pm1.2)$ & $\bf86.4 (\pm1.2)$ \\
\hline
\emph{\textbf{Ours w/o BN}} \\
\hline
E2E (1x64, CHT) & ResNet-FrozenBN & LSTM & 1 & $90.0(\pm7.5)$ & $77.3(\pm8.5)$ & $89.0 (\pm0.3)$ & $69.4 (\pm1.1)$ & $\bf85.5 (\pm0.6)$ \\
Partial freezing (1x256, CHT) &  &  & 1 & $\bf92.5(\pm5.9)$ & $\bf81.2(\pm9.6)$ & $\bf91.5 (\pm0.5)$ & $74.2 (\pm1.2)$ & $\bf88.0 (\pm0.8)$ \\
\hline
E2E (1x64, CHT) & ResNet-GN & LSTM & 1 & $91.4(\pm7.8)$ & $78.9(\pm9.7)$ & $90.3 (\pm1.3)$ & $71.5 (\pm3.0)$ & $\bf86.0 (\pm1.3)$ \\
Partial freezing (1x256, CHT) &  &  & 1 & $\bf{93.1}(\pm5.4)$ & $\bf{80.2}(\pm11.7)$ & $\bf92.1 (\pm0.8)$ & $73.9 (\pm1.9)$ & $\bf86.4 (\pm0.9)$ \\
\hline
E2E (1x64, CHT) & ConvNeXt & LSTM & 1 & $\bf{92.4}(\pm6.9)$ & $\bf{82.0}(\pm7.9)$ & $91.3 (\pm0.2)$ & $74.6 (\pm0.6)$ & $\bf87.4 (\pm0.7)$ \\
Partial freezing (1x256, CHT) &  &  & 1 & $\bf{93.5}(\pm6.5)$ & $\bf{82.9}(\pm10.1)$ & $\bf92.4 (\pm0.3)$ & $\bf76.9 (\pm1.0)$ & $\bf87.9 (\pm1.3)$ \\
\hline
\hline
\emph{Concurrent work}\\
\hline
LoViT~\citep{liu2023lovit} & Transf. & Transf. & 2 & $\it92.4 (\pm6.3)$ & $\it81.2 (\pm9.1)$ & $\it91.5$ & $\it74.2$ \\
SKiT~\citep{liu2023skit} & Transf. & Transf. & 2 & $\it93.4 (\pm5.2)$ & $\it82.6$ & $\it92.5$ & $\it76.7$ \\
\hline
\multicolumn{5}{l}{$\dagger$\emph{reproduced}}
\end{tabular}
}
\end{center}
\end{table*}

\begin{table}[t]
\begin{center}
\caption{\color{\colorrevtwo}State of the art in surgical phase recognition on AutoLaparo~\citep{wang2022autolaparo} with 10/4/7 split. We report mean$\pm$std.\ over runs.}
\label{tbl:autolaparo}
\resizebox{\linewidth}{!}{
\begin{tabular}{lcccc}
\hline
\multicolumn{5}{c}{\textbf{AutoLaparo}} \\
\hline
Method & Accuracy & Precision & Recall & Jaccard \\
\hline
SV-RCNet~\citep{jin2017sv} & $75.6$ & $64.0$ & $59.7$ & $47.2$ \\
TMRNet~\citep{jin2021temporal} & $78.2$ & $66.0$ & $61.5$ & $49.6$ \\
TeCNO~\citep{czempiel2020tecno} & $77.3$ & $66.9$ & $64.6$ & $50.7$ \\
Trans-SVNet~\citep{gao2021trans} & $78.3$ & $64.2$ & $62.1$ & $50.7$ \\
LoViT~\citep{liu2023lovit} & $81.4$ & $\bf85.1$ & $65.9$ & $56.0$ \\
SKiT~\citep{liu2023skit} & $82.9$ & $81.8$ & $70.1$ & $59.9$ \\
\hline
\emph{\textbf{ResNet-BN-LSTM}} \\
\hline
E2E (4x16, CHE) & $82.3 (\pm0.5)$ & $71.9 (\pm1.1)$ & $67.8 (\pm0.8)$ & $57.7 (\pm1.2)$ \\
Partial freezing (4x64, CHE) & $\bf85.2 (\pm2.3)$ & $78.0 (\pm2.4)$ & $\bf73.3 (\pm4.7)$ & $\bf64.1 (\pm5.7)$ \\
Partial freezing (1x256, CHT) & \color{red}$49.4 (\pm12.8)$ & \color{red}$54.0 (\pm3.7)$ & \color{red}$42.1 (\pm13.4)$ & \color{red}$26.7 (\pm10.0)$ \\
\hline
\emph{\textbf{ResNet-GN-LSTM}} \\
\hline
E2E (1x64, CHT) & $80.3 (\pm1.8)$ & $72.4 (\pm3.0)$ & $65.5 (\pm2.7)$ & $55.6 (\pm3.1)$ \\
Partial freezing (4x64, CHE) & $82.7 (\pm1.3)$ & $73.5 (\pm0.6)$ & $68.1 (\pm2.1)$ & $58.4 (\pm2.4)$ \\
Partial freezing (1x256, CHT) & $\bf85.7 (\pm0.8)$ & $73.7 (\pm2.0)$ & $\bf71.5 (\pm1.3)$ & $\bf63.0 (\pm1.3)$ \\
\hline
\emph{\textbf{ConvNeXt-LSTM}} \\
\hline
E2E (1x64, CHT) & $\bf83.0 (\pm1.2)$ & $73.8 (\pm0.5)$ & $66.8 (\pm1.4)$ & $57.7 (\pm1.4)$ \\
Partial freezing (4x64, CHE) & $82.8 (\pm0.8)$ & $79.3 (\pm4.7)$ & $66.3 (\pm0.8)$ & $57.5 (\pm1.1)$ \\
Partial freezing (1x256, CHT) & $\bf86.8 (\pm1.5)$ & $78.2 (\pm4.4)$ & $\bf72.0 (\pm1.9)$ & $\bf64.2 (\pm2.2)$ \\
\hline
\end{tabular}
}
\end{center}
\end{table}

\begin{table}[t]
\begin{center}
\caption{\color{\colorrevtwo}State of the art in surgical step recognition on CATARACTS~\citep{zisimopoulos2018deepphase} with a 25/25 split. Our models are trained on a 20/5/25 split to allow for model selection. State-of-the-art scores are printed bold and we underline scores which outperform all methods except \citet{chen2022spatio}.}
\label{tbl:cataracts}
\resizebox{\linewidth}{!}{
\begin{tabular}{lcccc}
\hline
\multicolumn{5}{c}{\textbf{CATARACTS}} \\
\hline
Method & Accuracy & Precision & Recall & Jaccard \\
\hline
3D-CNN~\citep{funke2019using} & $80.1$ & $66.2$ & $55.7$ & $45.9$ \\
SV-RCNet~\citep{jin2017sv} & $81.3$ & $66.0$ & $57.0$ & $47.2$ \\
TeCNO~\citep{czempiel2020tecno} & $79.0$ & $62.6$ & $56.9$ & $45.1$ \\
Trans-SVNet~\citep{gao2021trans} & $77.8$ & $61.3$ & $55.0$ & $43.8$ \\
Dual Pyramid~\citep{chen2022spatio} & $\bf84.2$ & $69.3$ & $\bf66.4$ & $\bf53.7$ \\
\hline
\emph{\textbf{ResNet-BN-LSTM}} \\
\hline
E2E (4x16, CHE) & $79.3 (\pm1.4)$ & $63.0 (\pm2.6)$ & $55.7 (\pm2.4)$ & $44.8 (\pm2.7)$ \\
E2E (1x64, CHT) & \color{red}$40.4 (\pm5.0)$ & \color{red}$35.4 (\pm8.8)$ & \color{red}$20.5 (\pm6.2)$ & \color{red}$11.8 (\pm4.3)$ \\
Partial freezing (4x64, CHE) & $79.9 (\pm1.7)$ & $60.4 (\pm2.7)$ & $55.9 (\pm0.4)$ & $43.7 (\pm0.9)$ \\
\hline
\emph{\textbf{ResNet-GN-LSTM}} \\
\hline
E2E (4x16, CHE) & $80.8 (\pm1.4)$ & \underline{\smash{$67.5 (\pm0.9)$}} & \underline{\smash{$57.3 (\pm2.3)$}} & $47.1 (\pm2.9)$ \\
E2E (1x64, CHT) & \underline{\smash{$82.1 (\pm0.6)$}} & $64.0 (\pm2.1)$ & \underline{\smash{$58.4 (\pm1.1)$}} & \underline{\smash{$47.5 (\pm0.9)$}} \\
Partial freezing (1x256, CHT) & $80.4 (\pm2.4)$ & $66.1 (\pm5.4)$ & \underline{\smash{$58.9 (\pm4.8)$}} & $47.2 (\pm5.1)$ \\
\hline
\emph{\textbf{ConvNeXt-LSTM}} \\
\hline
E2E (4x16, CHE) & \underline{\smash{$83.0 (\pm1.2)$}} & $\bf69.9 (\pm2.3)$ & \underline{\smash{$59.7 (\pm1.1)$}} & \underline{\smash{$49.3 (\pm0.6)$}} \\
E2E (1x64, CHT) & \underline{\smash{$83.8 (\pm0.8)$}} & $63.7 (\pm5.7)$ & \underline{\smash{$60.9 (\pm2.7)$}} & \underline{\smash{$49.7 (\pm3.6)$}} \\
Partial freezing (1x256, CHT) & \underline{\smash{$83.3 (\pm0.5)$}} & \underline{\smash{$66.8 (\pm3.5)$}} & \underline{\smash{$61.8 (\pm1.7)$}} & \underline{\smash{$50.3 (\pm0.5)$}} \\
\hline
\end{tabular}
}
\end{center}
\end{table}

{\bf Hypothesis 7: Simple end-to-end CNN-LSTMs achieve state-of-the-art performance.}
Finally, we compare our proposed approaches (end-to-end and partial freezing) to the state of the art in phase and {\color{\colorrevtwo}step recognition on three datasets}\footnote{\color{\colorrevtwo}The unconventional classification of SKiT/LoViT as concurrent on Cholec80 and previous work on AutoLaparo is a result of AutoLaparo experiments being added to this paper at a later time point.}.

{\color{\colorrevtwo}\textbf{Cholec80:}} To be comparable to previous work~\citep{gao2021trans,jin2017sv,jin2021temporal}, we retrain our models on a 32/8/40 split. As evaluation metrics, we report the mean video-wise accuracy and the mean video-wise balanced accuracy (i.e. mean video-wise macro recall) as well as variants of accuracy and Jaccard scores with relaxed boundaries, i.e.\ where errors at phase transitions are ignored under certain conditions. Relaxed metrics were proposed in the \emph{M2CAI 2016 Challenge}~\citep{stauder2016tum,twinanda2016endonet}.
Table~\ref{table:sota} shows our results.

Through partial freezing on 256-frame sequences, all our models with BN-free backbones (FrozenBN, GN and ConvNeXt) outperform the previous state of the art, including complex multi-stage methods with attention-based temporal models~\citep{gao2021trans,jin2021temporal,chen2022spatio}, complex augmentation pipelines~\citep{yi2022not} or Transformer backbones~\citep{chen2022spatio,he2022empirical}.
Models trained completely end to end (E2E) on shorter 64-frame sequences perform slightly worse. However, the ConvNeXt variant still achieves SOTA performance and the GN variant beats or matches other ResNet-based methods. Note that Not-E2E~\citep{yi2022not} achieves slightly higher relaxed accuracy but much lower relaxed Jaccard. The completely end-to-end FrozenBN model underperforms, as already observed in \emph{Hypothesis 3}.

While the newer ConvNeXt backbone is not necessarily comparable to ResNet backbones, note that our proposed strategies are necessary
to achieve strong performance. Specifically, ConvNeXt models do \emph{not} perform better than its ResNet variants with the \emph{CHE} strategy (Fig.~\ref{fig:carry_hidden}) or in 2-stage settings (Fig.~\ref{fig:e2e_2stage} \& Table~\ref{table:acc}). These results further indicate that improving visual-feature learning in simple end-to-end approaches might be more effective than designing more complex temporal models on sub-optimal visual features.

Finally, BN-based end-to-end models perform poorly with SV-RCNet's sliding-window evaluation (SWE) but improve by simply carrying hidden states during inference (CHE) without changing the training procedure. This indicates that previous end-to-end approaches were evaluated poorly. By freezing backbone layers, BN-based CNN-LSTMs also achieve competitive results and match the 64-frame, end-to-end GN model. Performance of ConvNeXt or 256-frame GN and FrozenBN models is however not reached. This shows that, with careful design, BN can still be effective. Still, models without BatchNorm enable simpler and a wider range of training procedures with longer training sequences and ultimately perform better.

{\color{\colorrevtwo}\textbf{AutoLaparo:} {\color{\colorrevone}With AutoLaparo containing long videos like Cholec80, we can reproduce our main findings (Table~\ref{tbl:autolaparo}): Maximizing temporal context through the proposed strategies} (Partial freezing, 1x256, CHT) outperforms its ablations without single-sequence batches and CHT (Partial freezing, 4x64, CHE) or partial freezing (E2E 1x64, CHT). These models even outperform the state of the art by a margin, including very recent, fully transformer-based approaches LoViT and SKiT. Although BatchNorm models perform very well with partial freezing, employing single-sequence batches and CHT, which works best with all other backbones, expectedly fails with BatchNorm.}

\begin{figure}[t]
    \centering
    \includegraphics[trim=0cm 12cm 0cm 0.2cm,clip,width=\linewidth]{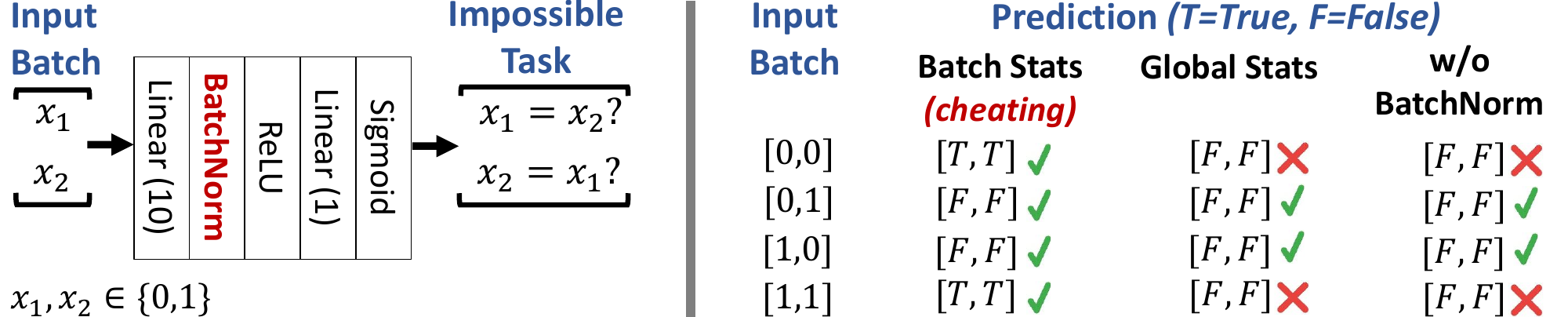}
    \caption{Impossible toy task to illustrate ``cheating'' with BN: \emph{predicting whether the input is equal to the other sample in the batch}. During training, batch stats can be used to solve the task but the model fails at test time using global stats.}
    \label{fig:toy}
\end{figure}

{\color{\colorrevtwo}\textbf{CATARACTS:} {\color{\colorrevone}Since CATARACTS contains shorter videos with more fine-grained annotations, the benefits of our context-maximizing strategies are expectedly smaller here. Nevertheless, we can still reproduce several findings (Table~\ref{tbl:cataracts}):} Single-sequence CHT (E2E, 1x64, CHT) outperforms multi-sequence batching (E2E, 4x16, CHE) with BN-free backbones but fails with BatchNorm. Partial freezing did not have a clear benefit and even underperformed with GN - likely due to fewer long-term dependencies. Despite not reaching SOTA performance (as the fully Transformer-based model by \citet{chen2022spatio} performs exceptionally well), our simple ConvNeXt and GN models consistently achieve the second highest scores, outperforming Transformer- and TCN-based multi-stage approaches (Trans-SVNet, TeCNO) as well as 3D-CNNs. BN-based models do not reach this level and again fail when using CHT, which was most effective for BN-free backbones.}

\subsection{Anticipation of Instrument Usage} \label{sec:anticipation}

{\bf Hypothesis 8: BatchNorm enables ``cheating'' in certain tasks.}
BatchNorm can leak information within batches to ``cheat'' certain objectives. While previous work finds empirical evidence for this effect through degrading performance~\citep{wu2021rethinking}, we conclusively demonstrate it by carefully designing a toy task: We propose an impossible task which cannot be solved without ``cheating'': \emph{Given training batches of size 2, predict for each input sample whether it is equal to the other sample in the batch}. We use a small neural network with BatchNorm (Fig.~\ref{fig:toy}, left) and with single binary inputs $x_i \in\{0,1\}$. Although this task is clearly unsolvable when each sample in the batch is processed individually, the BN model ``cheats'' by making predictions based on batch statistics (Fig.~\ref{fig:toy}, right). Models expectedly fail (always predict $False$) when using global statistics during inference or networks without BN.

{\bf Hypothesis 9: ``Cheating'' can cause BN models to fail in instrument anticipation.}
To visualize how BN can ``cheat'' during training in anticipation tasks, we evaluate BN in training mode, i.e.\ processing test videos offline in batches with sequence lengths equal to the training lengths and use batch (instead of global) statistics for normalization. The effect was most pronounced with sequences similar to the anticipation horizon (5min), so we opt for partial freezing with training sequences of 256 frames (4min16s).

\begin{figure}[t]
    \centering
    \begin{subfigure}[b]{0.475\linewidth}
        \centering
        \includegraphics[trim=.2cm 5.7cm 13.9cm .1cm,clip,width=\linewidth]{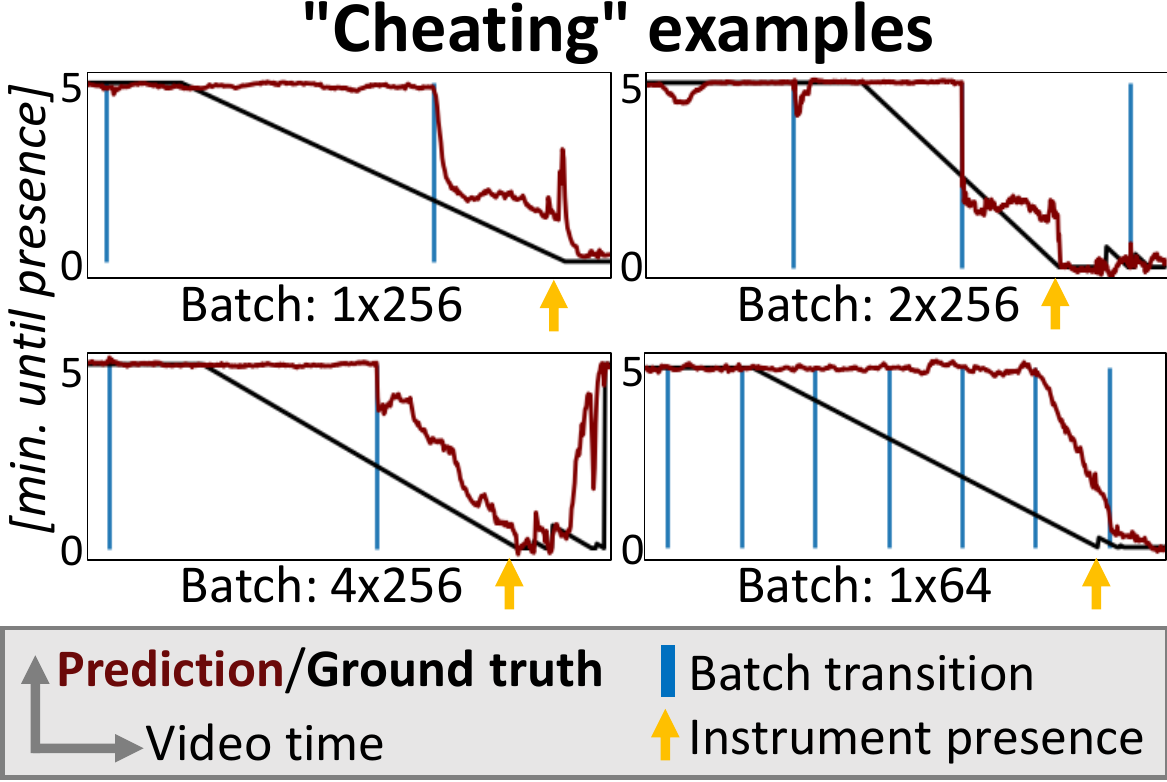}
        \caption{\emph{Cheating examples}}
        \label{fig:cheating_example}
    \end{subfigure}
    \hfill
    \begin{subfigure}[b]{0.5\linewidth}
        \centering
        \includegraphics[trim=.1cm .1cm 1.5cm .3cm,clip,width=\linewidth]{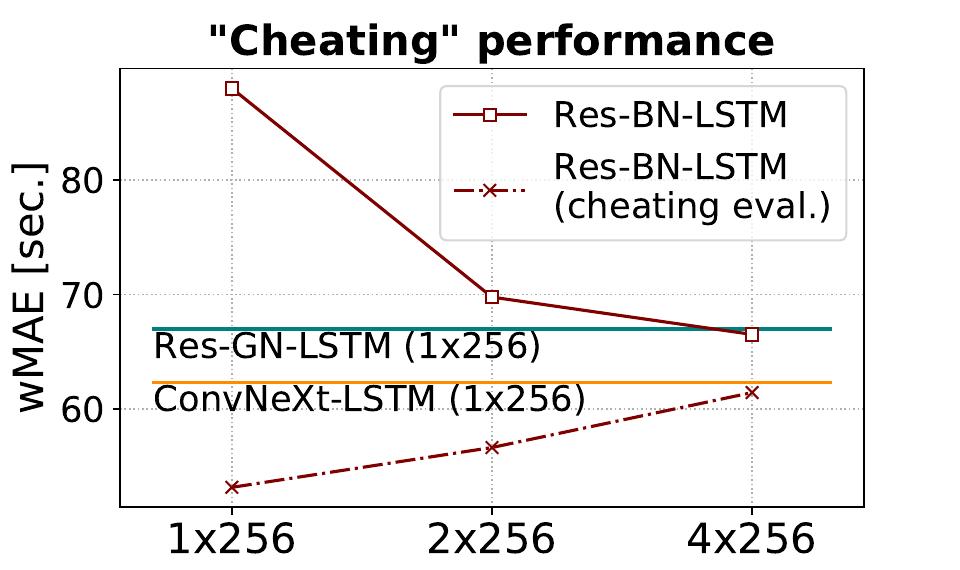}
        \caption{\emph{Cheating impairs BN models}}
        \label{fig:cheating_scores}
    \end{subfigure}
    \caption{\emph{Anticipation (Cholec80):} (a) During training, BN models ``cheat'' by detecting instruments in later frames through batch stats. We visualize this by using batch stats at test time. ``Cheating'' still happens in batches with multiple 
    or short 
    sequences. (b) By using batch stats at test time, BN models achieve overly good scores \emph{(cheating eval.)} but fail in a fair evaluation. Sampling multiple sequences per batch reduces but does not eliminate the effect.}
    \label{fig:cheating_anticipation}
\end{figure}

Fig.~\ref{fig:cheating_example} shows how anticipation predictions immediately improve in batches which contain an instrument occurrence at some point later in the batch. These examples visualize the ``cheating'' effect explicitly, not only indirectly through impact on  performance. Likely, the model has learned to recognize the instrument and the change in feature activations
becomes visible to previous frames through batch statistics. 
Namely, a sudden decrease of activations related to an instrument indicates that a higher mean was subtracted during normalization, so the instrument's occurrence later in the batch can be inferred. Interestingly, this effect is still visible (but less pronounced) in models trained on batches that contain multiple ($4\times256$) or shorter ($1\times64$) sequences.

In Fig.~\ref{fig:cheating_scores}, we investigate how the ``cheating'' phenomenon affects anticipation performance. BN models are evaluated either using batch statistics (\emph{cheating eval.}) or using global statistics, which would be the correct approach as future information is generally not available in an online anticipation setting. By using batch statistics at test time, the ``cheating'' model strongly (but unfairly) outperforms even BN-free end-to-end models. However, we can see that a stronger ``cheating'' effect (i.e.\ a lower error with \emph{cheating eval.}, dotted line) leads to a higher error during correct evaluation (solid line). By sampling multiple sequences per batch ($4\times256$), this effect is reduced and the BN model matches the performance of its GN counterpart. However, the gap between the two curves suggests that the model is still learning subpar features due to ``cheating''.

{\bf Hypothesis 10: End-to-end models outperform state of the art in anticipation.}
Finally, we compare our models and strategies to previous work in instrument anticipation~\citep{rivoir2020rethinking,yuan2021surgical}.
In Table~\ref{table:anticipation}, we observe that ConvNeXt and FrozenBN outperform previous work. The GN model only performs well when trained completely end-to-end but gives poor results with partial CNN freezing. As expected, BN models completely fail with single-sequence batches ($1\times64, 1\times256$) but also provide subpar performance with multiple long sequences ($2\times256, 4\times256, 4\times64$). Likely, this is in part caused by BN's ``cheating'' (see Fig.~\ref{fig:cheating_example}). However, other factors may have contributed since the partially frozen GN model underperforms as well. The end-to-end BN model trained on shorter 16-frame sequences performs surprisingly well. Likely, 16-second batches are too short for effective ``cheating''.

Across backbones, complete end-to-end training is more effective than freezing layers, possibly because anticipation is a more challenging task and requires more finetuning of CNN weights. Also, the CHE strategy outperforms CHT in this task. Presumably, the correlation between subsequent batches (and thus SGD steps) is not ideal for this more difficult task. Similarly, FrozenBN models underperformed in the CHT setting in phase recognition (Fig. \ref{fig:carry_hidden}, \ref{fig:other}). Nevertheless, we provide strong evidence for the effectiveness of end-to-end learning also for anticipation and show that BN models are limited to short sequences and require careful engineering due to the ``cheating'' phenomenon.

\begin{table}[t]
\begin{center}
\caption{Comparison of anticipation methods (Cholec80). Scores (mean+std. over runs) which outperform state of the art are highlighted (\textbf{bold})}
\label{table:anticipation}
\resizebox{\linewidth}{!}{
\begin{tabular}{llcccc}
\hline
\multicolumn{6}{c}{\textbf{Anticipation error $\downarrow$ (wMAE in sec.)}} \\
\hline
SOTA & \multicolumn{4}{l}{End-to-end AlexNet-LSTM~\citep{rivoir2020rethinking}} & $66.02 (\pm0.85)$ \\
& \multicolumn{4}{l}{2-stage ResNet-TCN~\citep{yuan2021surgical} (variant with only visual features)} & $65.34 (\pm0.59)$ \\
\hline
\multicolumn{2}{l}{\textbf{CNN-LSTM:}} & \textbf{Res-BN} & \textbf{Res-FrozenBN} & \textbf{Res-GN} & \textbf{ConvNeXt} \\
\hline
E2E & 1x64, CHT & - & $66.36 (\pm1.30)$ & $67.18 (\pm1.39)$ & $\bf64.74 (\pm0.12)$ \\
& 1x64, CHE & $\color{red}83.78 (\pm3.29)$ & $\bf62.10 (\pm0.06)$ & $\bf64.36 (\pm0.28)$ & $\bf61.72 (\pm0.28)$ \\
& 4x16, CHE & $\bf62.06 (\pm0.95)$ & - & - & - \\
\hline
Partial & 1x256, CHE & $\color{red}88.02 (\pm16.58)$ & $\bf63.10 (\pm0.31)$ & $67.00 (\pm0.87)$ & $\bf62.34 (\pm0.33)$ \\
freezing & 2x256, CHE & $69.78 (\pm0.69)$ & - & - & - \\
& 4x256, CHE & $66.52 (\pm0.51)$ & - & - & - \\
& 4x64, CHE & $65.30 (\pm0.50)$ & - & - & - \\
\hline
\end{tabular}
}
\end{center}
\end{table}

\section{Limitations and Scope}
{\color{\colorrevone}Our proposed strategies aim at increasing temporal context, and thus seem mostly applicable to tasks where this is relevant, e.g. long videos and workflow-based tasks such as cooking or other complex activities. Its impact on other video tasks remains open.}
Further, our experiments focus on 2D backbones, due to limitations in the surgical domain, and LSTMs, to show the effectiveness of simple, state-based end-to-end models for online tasks.
While the incompatibility of BN and single-sequence batches is general to the video setting, investigating the feasibility and effectiveness of end-to-end approaches for other domains, backbones and temporal models requires further work and falls outside the scope of our study.
\citet{liu2022empirical} recently provided an extensive, insightful study on end-to-end temporal action detection with a wider range of models. However, BN issues are not discussed, although the common FrozenBN solution is utilized. Our work, on the other hand, aims at providing a deeper understanding of BN's limitations in long-video settings but restricts practical implications to a smaller range of models.

{\color{\colorrevone}Nevertheless, we provide some hypotheses and ideas for future directions regarding other architectures:
\emph{(a)} 3D-CNNs do not differ in their use of normalization layers and thus BN issues and its solutions would likely be similar. Yet, available, non-causal 3D-CNNs pose practical limitations for online tasks since popular sliding-window approaches would pose strong memory constraints.
\emph{(b)} 2D Vision-Transformer backbones could be used in place of BN-free CNNs since they typically do not contain BatchNorm. Thus, we believe the increasing popularity of Transformers in computer vision represents an opportunity for end-to-end video learning.
\emph{(c)} For temporal modeling, CHT seemed to be an important component for pushing performance past SOTA models in our experiments and this strategy is only directly applicable in RNNs. However, for other temporal models like TCNs or Transformers, similar strategies could be conceivable by carrying features of convolutional or attention layers across batches. This could be an interesting direction for future work.}

\section{Conclusion}
We investigate the limitations of BatchNorm for end-to-end video learning, which are especially relevant in surgical workflow tasks where CNN backbones require finetuning. In a detailed literature review, we reveal how research has circumvented these problems by moving towards complex, multi-stage approaches.
We identify situations where BN models fail and how it affects end-to-end training strategies in simple CNN-LSTMs. Specifically, BN restricts the length of training sequences. In online settings, can interfere with hidden-state handling of LSTMs as well as leak information from future frames. We show that the latter can be used by models to ``cheat'' in anticipation tasks. We also find that the common ad-hoc solution FrozenBN is only effective in restricted settings. Instead, even simple CNN-LSTMs achieve state-of-the-art performance in {\color{\colorrevtwo}three surgical workflow tasks} by avoiding BatchNorm entirely {\color{\colorrevone}and thus enabling longer temporal contexts.}

We believe a comprehensive and deeper understanding of BN's limitations in video tasks is crucial for future research in surgical workflow analysis.
End-to-end learning with simple models appears favorable over complex, multi-stage approaches but we hypothesize that BN issues have silently hidden its advantages. Moving forward, the field could benefit from reconsidering end-to-end approaches and further investigating BN-free backbones. In natural-video tasks like action segmentation, where end-to-end learning currently plays only a minor role, re-evaluating training strategies could potentially be valuable as well.

\section*{Acknowledgments} Funded by the German Research Foundation (DFG, Deutsche Forschungsgemeinschaft) as part of Germany’s Excellence Strategy – EXC 2050/1 – Project ID 390696704 – Cluster of Excellence “Centre for Tactile Internet with Human-in-the-Loop” (CeTI) of Technische Universität Dresden.

\bibliographystyle{model2-names.bst}\biboptions{authoryear}
\bibliography{medima-template.bbl}

\newpage

\appendix

\section{Results for Surgical Phase Recognition}
Tables \ref{table:acc} and \ref{table:f1} provide an overview of accuracy and F1 scores for surgical phase recognition. The accuracy scores were presented in the form of figures in the main paper (Fig. \ref{fig:e2e_2stage} \& \ref{fig:train_strategies}).

In Table~\ref{table:lr}, we provide additional evidence that models with FrozenBN exhibit more instability during training.

\begin{table}[t]
\begin{center}
\caption{Accuracy scores (mean + sample std. over runs) for surgical phase recognition on TeCNO's~\citep{czempiel2020tecno} split (40 train. videos, 8 val., 32 test) used for our main analysis. Per run, accuracy is computed per video and then averaged. We \textbf{highlight} scores which outperform all 2-stage approaches by at least its sample standard deviation. Note that these results were presented in the form of figures in the main paper}
\label{table:acc}
\resizebox{\linewidth}{!}{
\begin{tabular}{lccccc}
\hline
\multicolumn{5}{c}{\textbf{Accuracy (video-wise) $\uparrow$}} & \textbf{Fig.} \\
\hline
& BN & FrozenBN & GN & ConvNeXt \\
\hline
\textbf{2-Stage} (CNN-...) \\
\hline
LSTM & $89.1 (\pm0.5)$ & - & $87.2 (\pm0.7)$ & $87.8 (\pm0.3)$ & \multirow{3}{*}{\ref{fig:e2e_2stage}}\\
TCN \emph{(*TeCNO reimpl.)} & *$89.4 (\pm0.8)$ & - & $88.4 (\pm0.8)$ & $88.2 (\pm0.1)$ \\
TeCNO~\citep{czempiel2020tecno} \emph{(original)} & $88.56 (\pm0.27)$ & - & - & - \\
\hline
\textbf{End-to-end} (CNN-LSTM) \\
\hline
$8\times8$, sliding eval. & $85.9 (\pm0.1)$ & $85.2 (\pm0.6)$ & $83.0 (\pm2.2)$ & $85.9 (\pm0.2)$ & \multirow{4}{*}{\ref{fig:seq_len}} \\
$4\times16$, sliding eval. & $86.5 (\pm1.4)$ & $86.1 (\pm1.2)$ & $85.8 (\pm0.7)$ & $88.0 (\pm0.9)$ \\
$2\times32$, sliding eval. & $86.5 (\pm0.4)$ & $86.5 (\pm1.2)$ & $87.5 (\pm2.3)$ & $87.9 (\pm0.5)$ \\
$1\times64$, sliding eval. & $\color{red}57.2 (\pm4.2)$ & $88.4 (\pm1.9)$ & $88.4 (\pm0.8)$ & $89.3 (\pm1.2)$ \\
\hline
$8\times8$, carry-hidden eval. & $89.4 (\pm0.4)$ & $89.0 (\pm0.4)$ & $86.8 (\pm1.3)$ & $88.1 (\pm0.2)$ & \multirow{4}{*}{\ref{fig:carry_hidden}} \\
$4\times16$, carry-hidden eval. & $89.6 (\pm1.0)$ & $89.4 (\pm1.0)$ & $87.9 (\pm0.8)$ & $89.1 (\pm1.1)$ \\
$2\times32$, carry-hidden eval. & $88.2 (\pm0.7)$ & $89.4 (\pm1.0)$ & $88.9 (\pm2.3)$ & $88.6 (\pm0.6)$ \\
$1\times64$, carry-hidden eval. & $\color{red}59.7 (\pm3.8)$ & $89.6 (\pm1.1)$ & $89.3 (\pm0.4)$ & $89.5 (\pm1.1)$ \\
\hline
$1\times64$, carry-hidden train. & $\color{red}55.6 (\pm3.1)$ & $89.4 (\pm0.4)$ & $\bf90.5 (\pm0.9)$ & $\bf92.3 (\pm1.5)$ & \ref{fig:e2e_2stage} \& \ref{fig:carry_hidden} \\
\hline
\textbf{Partially frozen} (CNN-LSTM) \\
\hline
$4\times64$, carry-hidden eval. & $90.1 (\pm2.2)$ & - & - & - & \multirow{3}{*}{\ref{fig:frozen}}\\
$2\times128$, carry-hidden eval. & $88.5 (\pm1.6)$ & - & - & - \\
$1\times256$, carry-hidden eval. & $\color{red}76.7 (\pm1.5)$ & $90.0 (\pm0.7)$ & $\bf91.6 (\pm0.5)$ & $90.6 (\pm2.6)$ \\
\hline
$1\times64$, carry-hidden train. & - & $\bf92.4 (\pm0.4)$ & $89.7 (\pm1.8)$ & $\bf93.3 (\pm0.3)$ & \multirow{3}{*}{\ref{fig:frozen}} \\
$1\times128$, carry-hidden train. & - & $\bf92.0 (\pm0.5)$ & $\bf92.3 (\pm0.5)$ & $\bf92.5 (\pm1.2)$ \\
$1\times256$, carry-hidden train. & - & $\bf92.1 (\pm0.4)$ & $\bf92.6 (\pm0.5)$ & $\bf92.6 (\pm0.1)$ \\
\hline
\end{tabular}
}
\end{center}
\end{table}

\begin{table}[t]
\begin{center}
\caption{F1 scores (mean + sample std. over runs) for surgical phase recognition on TeCNO's~\citep{czempiel2020tecno} split (40 train. videos, 8 val., 32 test) used for our main analysis. Per training run, we compute the frame-based confusion matrix over the entire test set, compute a single F1 score per phase and average over phases. This strategy deviates from video-wise approaches in previous work but is less ambiguous and gives smoother scores in edge cases. We \textbf{highlight} scores which outperform all 2-stage approaches by at least its sample standard deviation}
\label{table:f1}
\resizebox{\linewidth}{!}{
\begin{tabular}{lccccc}
\hline
\multicolumn{5}{c}{\textbf{F1 (frame-wise) $\uparrow$}} & \textbf{Fig.} \\
\hline
& BN & FrozenBN & GN & ConvNeXt \\
\hline
\textbf{2-Stage} \\
\hline
LSTM & $83.2 (\pm0.9)$ & - & $80.2 (\pm1.2)$ & $81.0 (\pm0.2)$ & \multirow{2}{*}{\ref{fig:e2e_2stage}} \\
TCN & $83.4 (\pm1.2)$ & - & $81.0 (\pm1.0)$ & $81.7 (\pm0.6)$ \\
\hline
\textbf{End-to-end} (CNN-LSTM) \\
\hline
$8\times8$, sliding eval. & $79.3 (\pm0.9)$ & $78.8 (\pm0.2)$ & $74.6 (\pm2.1)$ & $78.4 (\pm1.2)$ & \multirow{4}{*}{\ref{fig:seq_len}} \\
$4\times16$, sliding eval. & $80.4 (\pm1.7)$ & $80.4 (\pm0.8)$ & $78.8 (\pm0.7)$ & $82.3 (\pm1.0)$ \\
$2\times32$, sliding eval. & $79.1 (\pm0.9)$ & $81.1 (\pm0.5)$ & $81.5 (\pm2.7)$ & $81.9 (\pm1.3)$ \\
$1\times64$, sliding eval. & $\color{red}36.3 (\pm3.1)$ & $83.2 (\pm1.4)$ & $81.8 (\pm0.7)$ & $84.0 (\pm1.0)$ \\
\hline
$8\times8$, carry-hidden eval. & $83.2 (\pm1.0)$ & $82.8 (\pm0.8)$ & $79.4 (\pm0.8)$ & $81.9 (\pm0.6)$ & \multirow{4}{*}{\ref{fig:carry_hidden}} \\
$4\times16$, carry-hidden eval. & $83.6 (\pm1.6)$ & $83.2 (\pm0.9)$ & $81.5 (\pm0.5)$ & $83.7 (\pm1.2)$ \\
$2\times32$, carry-hidden eval. & $81.3 (\pm0.9)$ & $82.8 (\pm1.1)$ & $82.9 (\pm2.9)$ & $83.2 (\pm0.8)$ \\
$1\times64$, carry-hidden eval. & $\color{red}37.1 (\pm3.0)$ & $82.8 (\pm1.9)$ & $82.9 (\pm0.5)$ & $\bf84.5 (\pm0.9)$ \\
\hline
$1\times64$, carry-hidden train. & $\color{red}40.0 (\pm4.1)$ & $82.2 (\pm0.2)$ & $83.9 (\pm0.7)$ & $\bf86.2 (\pm1.7)$ & \ref{fig:e2e_2stage} \& \ref{fig:carry_hidden} \\
\hline
\textbf{Partially frozen} (CNN-LSTM) \\
\hline
$4\times64$, carry-hidden eval. & $84.5 (\pm1.4)$ & - & - & - & \multirow{3}{*}{\ref{fig:frozen}}\\
$2\times128$, carry-hidden eval. & $79.8 (\pm2.6)$ & - & - & - \\
$1\times256$, carry-hidden eval. & $\color{red}58.8 (\pm0.9)$ & $82.9 (\pm2.2)$ & $83.7 (\pm0.9)$ & $\bf85.8 (\pm2.4)$ \\
\hline
$1\times64$, carry-hidden train. & - & $\bf85.3 (\pm1.3)$ & $82.4 (\pm1.0)$ & $\bf87.4 (\pm0.7)$ & \multirow{3}{*}{\ref{fig:frozen}} \\
$1\times128$, carry-hidden train. & - & $\bf85.0 (\pm0.5)$ & $83.8 (\pm1.7)$ & $\bf87.5 (\pm0.9)$ \\
$1\times256$, carry-hidden train. & - & $\bf85.5 (\pm0.6)$ & $\bf84.8 (\pm0.4)$ & $\bf87.1 (\pm0.5)$ \\
\hline
\end{tabular}
}
\end{center}
\end{table}

\begin{table}[t]
\begin{center}
\caption{FrozenBN models exhibit more instability during training than GN models. Initial learning rates of FrozenBN models often had to be reduced in order to prevent model collapse {\color{red}(red)}, but higher learning rates were preferable in cases without collapse.}
\label{table:lr}
\resizebox{.85\linewidth}{!}{
\begin{tabular}{l|cc|cc}
\hline
\textbf{Learning rate} & \multicolumn{4}{c}{\textbf{Accuracy (video-wise) $\uparrow$}} \\
\hline
& \multicolumn{2}{c|}{\textbf{FrozenBN}} & \multicolumn{2}{c}{\textbf{GN}} \\
\hline
& \multicolumn{4}{c}{E2E (1x64)} \\
\hline
& CHE & CHT & CHE & CHT \\
\hline
$10^{-6}$ & $88.2 (\pm0.1)$ & $\bf89.4 (\pm0.4)$ & $83.7 (\pm2.3)$ & $88.4 (\pm0.2)$ \\
$10^{-5}$ & $\bf89.6 (\pm1.1)$ & $\color{red}39.8 (\pm0.0)$ & $\bf89.3 (\pm0.4)$ & $\bf90.5 (\pm0.9)$ \\
\hline
\hline
& \multicolumn{4}{c}{Partial freezing (1x256)}\\
\hline
& CHE & CHT & CHE & CHT \\
\hline
$10^{-5}$ & $\bf90.0 (\pm0.7)$ & $\bf92.1 (\pm0.4)$ & $86.9 (\pm0.5)$ & $89.5 (\pm0.7)$ \\
$10^{-4}$ & $\color{red}39.8 (\pm0.0)$ & $\color{red}39.8 (\pm0.0)$ & $\bf91.1 (\pm0.9)$ & $\bf92.6 (\pm0.5)$ \\
\hline
\end{tabular}
}
\end{center}
\end{table}

\section{Results for Online Action Segmentation}
Tables \ref{table:gtea} and \ref{table:50salads} show our results on the GTEA~\citep{fathi2011learning} and 50Salads~\citep{stein2013combining} datasets. These are the same results which were presented as bar plots in Fig.~\ref{fig:other} of the paper. Differently from most previous work, we formulate action segmentation as an online task on these datasets. We use the suggested splits for K-fold cross validation but for each training run, we use one set for validation (model selection), one set for testing and the remaining for training. We use the same hyperparameters as for surgical phase recognition (see Sec.~\ref{sec:hyp}).

\begin{table}[t]
\begin{center}
\caption{Accuracy scores for \emph{online} action segmentation on the \emph{GTEA} dataset~\citep{fathi2011learning}. We use the suggested splits for 4-fold cross validation. Note that most works propose offline methods. For each backbone, we \textbf{highlight} the training strategy (CHE vs. CHT) which performs better. Note that these results were presented in the form of a bar plot in the main paper.}
\label{table:gtea}
\resizebox{.8\linewidth}{!}{
\begin{tabular}{lcccc}
\hline
\multicolumn{5}{c}{\textbf{GTEA}} \\
\hline
\multicolumn{5}{c}{\textbf{Accuracy (video-wise) $\uparrow$}} \\
\hline
& BN & FrozenBN & GN & ConvNeXt \\
\hline
\textbf{2-Stage} \\
\hline
LSTM & $60.1$ & - & - & - \\
TCN & $63.2$ & - & - & - \\
\hline
\textbf{End-to-end, $\bf1\times64$} \\
\hline
CHE & - & $\bf66.6$ & $63.9$ & $63.5$ \\
CHT & $\color{red}43.6$ & $64.3$ & $\bf66.8$ & $\bf68.9$ \\
\hline
\end{tabular}
}
\end{center}
\end{table}

\begin{table}[t]
\begin{center}
\caption{Accuracy scores for \emph{online} action segmentation on the \emph{50Salads} dataset~\citep{stein2013combining}. We use the suggested splits for 5-fold cross validation. Note that most works propose offline methods. For each backbone, we \textbf{highlight} the training strategy (CHE vs. CHT) which performs better. Note that these results were presented in the form of a bar plot in the main paper.}
\label{table:50salads}
\resizebox{.8\linewidth}{!}{
\begin{tabular}{lcccc}
\hline
\multicolumn{5}{c}{\textbf{50Salads}} \\
\hline
\multicolumn{5}{c}{\textbf{Accuracy (video-wise) $\uparrow$}} \\
\hline
& BN & FrozenBN & GN & ConvNeXt \\
\hline
\textbf{2-Stage} \\
\hline
LSTM & $69.4$ & - & - & - \\
TCN & $72.0$ & - & - & - \\
\hline
\textbf{End-to-end, $\bf1\times64$} \\
\hline
CHE & - & $\bf77.9$ & $80.4$ & $79.3$ \\
CHT & $\color{red}23.5$ & $77.5$ & $\bf83.8$ & $\bf83.3$ \\
\hline
\end{tabular}
}
\end{center}
\end{table}

\section{Model Architectures}
\begin{figure*}
    \centering
    \includegraphics[trim=0cm 1cm 0cm 3cm,clip,width=.7\textwidth]{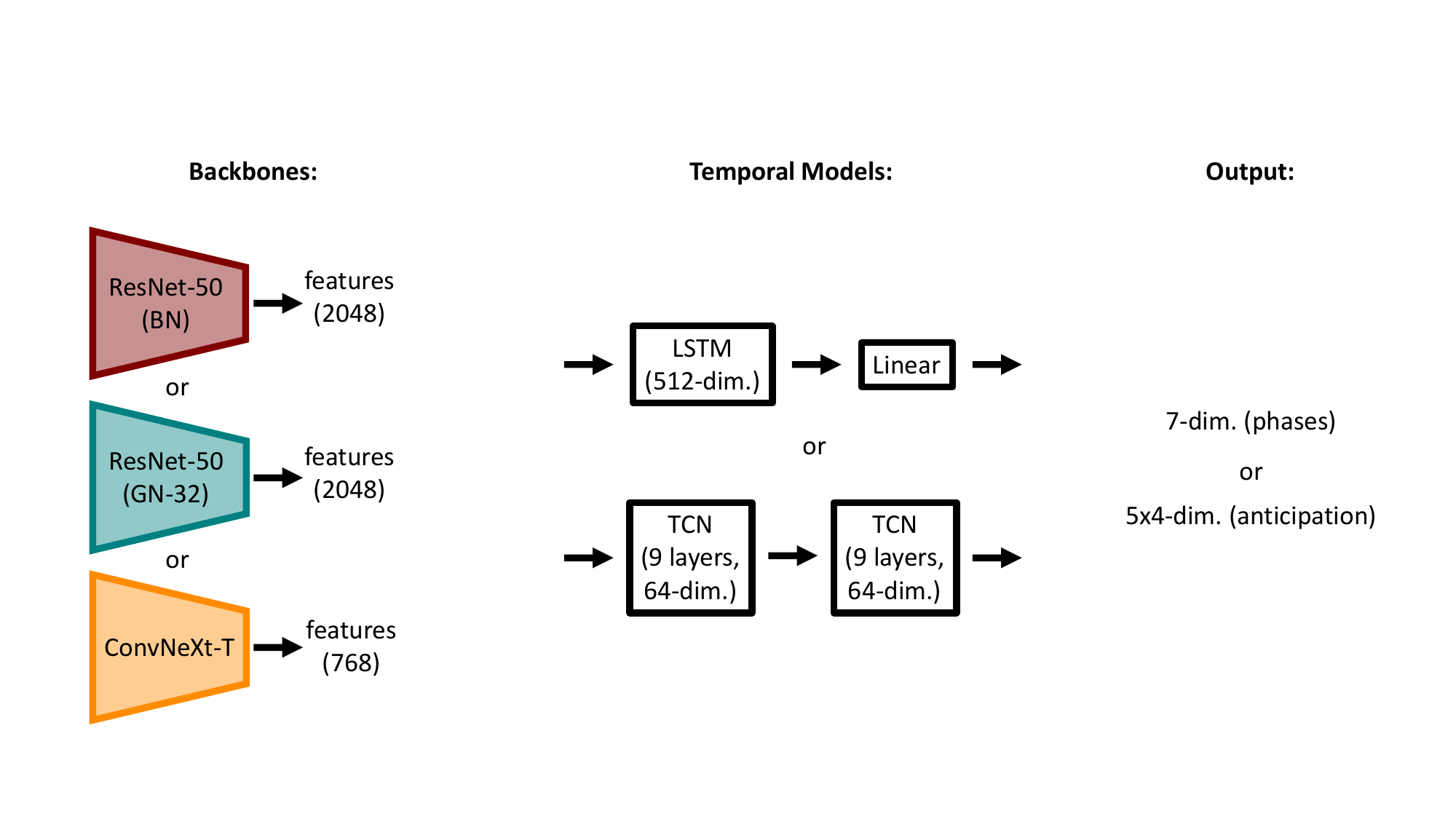}
    \caption{Overview of model architectures used in our experiments. For instrument anticipation, we follow the original task formulation~\citep{rivoir2020rethinking}. For each of 5 instruments, 4 prediction values are required (1 for duration prediction and 3 for the auxiliary classification task)}
    \label{fig:arch}
\end{figure*}

Fig.~\ref{fig:arch} shows the network architectures used in our experiments. For the GroupNorm-based Resnet-50, we use GroupNorm with 32 groups since this performed best in the original paper~\citep{wu2018group} and has a pretrained model available\footnote{\url{https://github.com/ppwwyyxx/GroupNorm-reproduce/releases/tag/v0.1}}. For 2-stage TCN models, we follow TeCNO~\citep{czempiel2020tecno} as close as possible. Although not reported in the paper, we follow their exact model configuration with 2 stages, 9 layers per stage and 64 feature channels per layer\footnote{\url{https://github.com/tobiascz/TeCNO/issues/6\#issuecomment-802851853}}.

\begin{figure*}
    \centering
    \includegraphics[trim=0cm 1.5cm 0cm 1.2cm,clip,width=.7\textwidth]{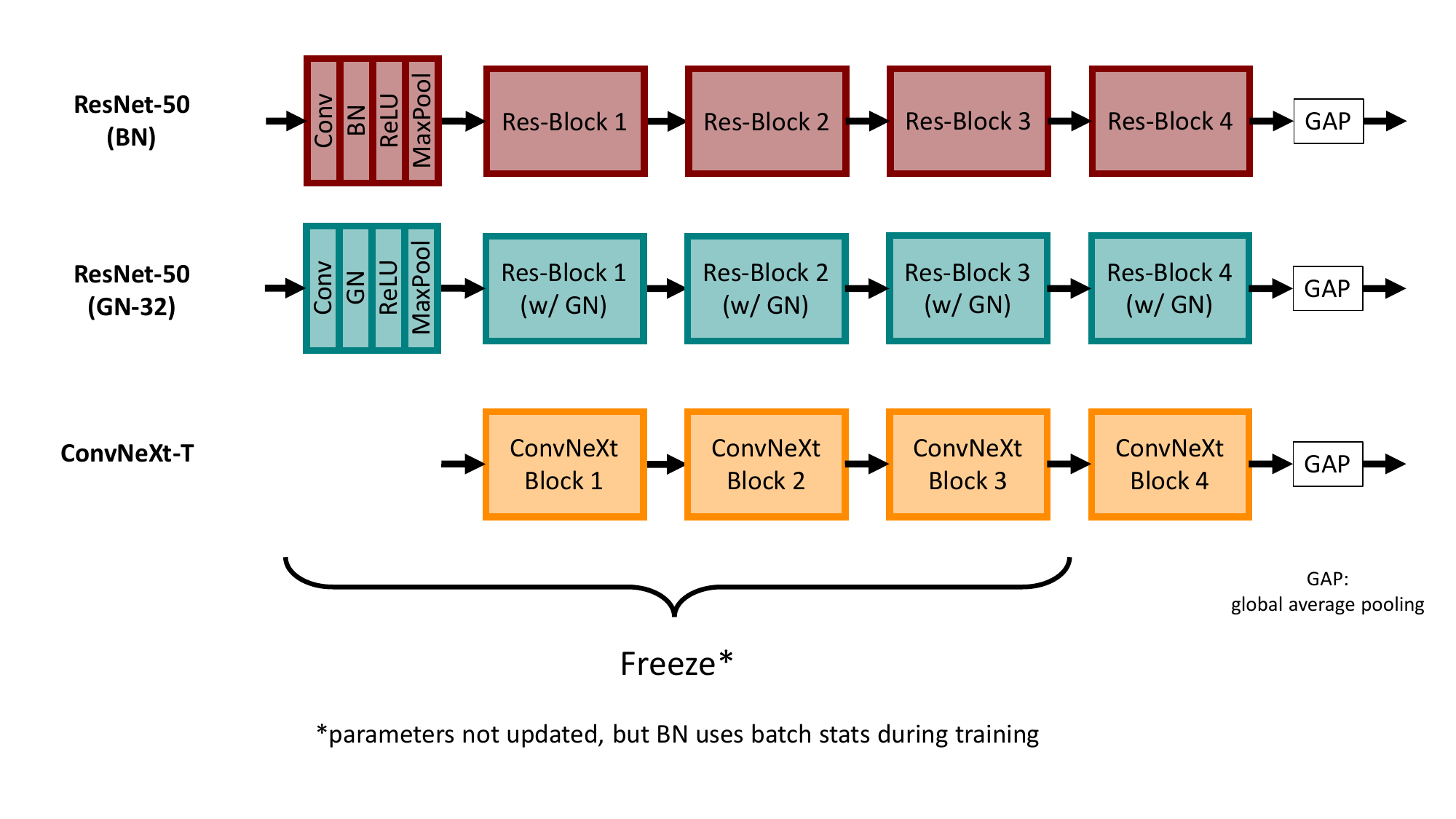}
    \caption{How backbones were frozen in the 'partially frozen' models. Note that with a BN-based backbone, batch statistics were still used for normalization during training. Setting BN to evaluation mode (i.e. using ImageNet's global stats) in frozen layers yielded similar results.}
    \label{fig:frozen_blocks}
\end{figure*}

Fig.~\ref{fig:frozen_blocks} shows which backbone blocks are frozen in the respective experiments. Note that with a BN-based backbone, batch statistics were still used for normalization during training. Setting BN to evaluation mode (i.e. using ImageNet's global stats) in frozen layers yielded similar results.

\section{Task Formulations}

\subsection{Surgical Phase Recognition}
Surgical phase recognition on the Cholec80 dataset~\citep{twinanda2016endonet} is a dense temporal segmentation task with 7 phases and is therefore a simple multi-class classification when viewed at the frame level. Given a frame $x$ with corresponding class label $c \in \{1,\dots,7\}$ and a softmax-activated neural network $f$ which outputs a $7$D vector per input image, we use the standard cross-entropy loss for optimization:
\begin{align}
    L_{phase}(x,c) = -\log f(x)[c]
\end{align}

\subsection{Anticipation of Surgical Instrument Usage}
For instrument ancitipation we follow the task formulation from previous work~\citep{rivoir2020rethinking}. We anticipate 5 different instruments on the Cholec80 dataset, where the main task is to predict the remaining duration until occurrence of each instrument at each time point up to a horizon of 5 minutes. As an auxiliary classification task, we also predict whether each instrument is present and whether or not it will appear within the next 5 minutes (i.e. inside or outside of horizon).

Given a frame $x$ and the corresponding remaining durations $t_i$ for each instrument $i \in \{1,\dots,5\}$ and the horizon $h=5$, the ground-truth value for the main regression task is defined as
\begin{align}
    y_i = \text{min}(t_i,h) \text{ .}
\end{align}
The corresponding ground-truth class for the auxiliary classification task is
\begin{align}
    c_i =
    \begin{cases}
        1, & y_i = 0 \text{ (instrument present)} \\
        2, & 0 < y_i < h \text{ (inside horizon)} \\
        3, & y_i = h \text{ (outside horizon)}
    \end{cases}
    \text{ .}
\end{align}
We then jointly optimize both tasks with the SmoothL1~\citep{yengera2018less} loss for regression and cross entropy for classification:
\begin{align}
    L_{ant}(x,y) = \sum_{i=1}^5 \text{SmoothL1}(f_{reg}^i(x),y_i) - \lambda \log f_{cls}^i(x)[c_i] \text{ ,}
\end{align}
where $f_{reg}^i$ and $f_{cls}^i$ are the regression and classification predictions of a neural network for instrument $i$ and $\lambda=0.01$ is a scaling factor for the auxiliary task.

\section{Training Details \& Hyperparameters} \label{sec:hyp}

Tables \ref{table:hyp_global}, \ref{table:hyp_varying} and \ref{table:hyp_dataset} summarize our training hyperparameters. Some of our choices might require clarification:
\begin{enumerate}
    \item In 'partially frozen' models for phase recognition, we trained for 100 epochs on the 40/8/32 split, but 200 epochs on the 32/8/40 split for state-of-the-art comparison due to the smaller training set. This was only done for efficiency reasons, since 100 epochs seemed to be enough with 40 train videos. The 'partial freezing' models did not deteriorate with longer training duration.
    \item When training only the backbone (for 2-stage approaches), we shortened the definition of an epoch, since the CNN overfit too quickly on the image task and would provide bad features for the temporal model in the subsequent stage.
    \item Despite our use of different learning rates, our decision rule was very simple: We use a default $10^{-5}$ for models where all parameters are trained (\emph{End-to-end} \& \emph{Backbone only}) and increased the learning by a factor of $10$ for shallower models (\emph{Partially frozen}). For training only the temporal models in 2-stage approaches (\emph{Temporal only}), we follow TeCNO~\citep{czempiel2020tecno} and use  $5\cdot10^{-4}$, which seemed to perform best in our experiments, too.
    \item Except for the number of epochs, we use exactly the same hyperparameters for phase recognition and anticipation. Convergence was much slower in anticipation.
\end{enumerate}

\begin{table}[t]
\begin{center}
\caption{Overview of global hyperparameters (used in all settings)}
\label{table:hyp_global}
\resizebox{.9\linewidth}{!}{
\begin{tabular}{l|c}
\hline
Optimizer & AdamW \\
Weight decay & $0.01$ \\
Image size & $216 \times 384$ \\
Data aug. & Shift, Scale, Crop, RGB shift, Brightness shift, Contrast shift \\
\hline
\end{tabular}
}
\end{center}
\end{table}

\begin{table}[t]
\begin{center}
\caption{Overview of hyperparameters which depend on the training strategy or task (for Cholec80).}
\label{table:hyp_varying}
\resizebox{\linewidth}{!}{
\begin{tabular}{l|c|c|c|c|c}
\hline
\textbf{Training type} & \multicolumn{2}{|c|}{\textbf{Epochs}} & \textbf{Learn. rate} & \textbf{Batch size} & \textbf{Batches per epoch} \\
\hline
& \multicolumn{1}{c|}{Phase} & \multicolumn{1}{c|}{Antic.} & \multicolumn{3}{c}{Both tasks} \\
\hline
Backbone only & 50 & 300 & $10^{-5}$ & 32 & \#frames / 256 \\
Temporal only & 100 & 100 & $5\cdot10^{-4}$ & whole video & \#videos \\
End-to-end & 200 & 300 & $10^{-5}\dagger$ & 64 & \#frames / batch size \\
Partially frozen & 200* & 300 & $10^{-4}\dagger$ & 64,128,256 & \#frames / batch size \\
\hline
\multicolumn{6}{l}{*100 on the 40/8/32 split. Solely for efficiency reasons. Longer training did not decrease performance.} \\
\multicolumn{6}{l}{\specialcell{$\dagger$FrozenBN models which collapsed under these parameters used a reduced learning rate by factor 10\\(E2E/CHT, frozen/CHE, frozen/CHT, on Cholec80 and 50Salads)}} \\
\end{tabular}
}
\end{center}
\end{table}

{\color{\colorrevtwo}
\subsection{Varying Hyperparameters on Other Datasets} \label{sec:hyp_other}

When retraining models on datasets other than Cholec80, we had to make changes to some hyperparameters:

\textbf{50Salads:} No changes were made.

\textbf{GTEA:} We did not have to reduce learning rates for FrozenBN models. All other hyperparameters are identical to Cholec80.

\textbf{AutoLaparo:} We increased the learning rate of partially frozen ConvNeXt models by a factor of $5$ (from $10^{-4}$ to $5\cdot10^{-4}$). In all other settings, this was not effective. Further, we found that accuracy was not a suitable metric for model selection on AutoLaparo due to the highly imbalanced dataset. Sometimes, models would achieve very high validation accuracy but low phase-based scores early during training and thus would lead to poor models being selected for testing. Instead, we used frame-based F1 for model selection.

\textbf{CATARACTS:} We increased learning rates of all ConvNeXt and BN models by a factor of 5 (E2E: $5\cdot10^{-5}$, partial freezing: $5\cdot10^{-4}$). For GN, this was not effective, so we kept the initial learning rates.
}

\begin{table}[t]
\begin{center}
\caption{Datasets}
\label{table:hyp_dataset}
\resizebox{.9\linewidth}{!}{
\begin{tabular}{l|cccc}
\hline
& FPS & Image resolution & Classes & Videos \\
\hline
Cholec80~\citep{twinanda2016endonet} & $1$ & $216\times384$ & $7$ & $80$ \\
AutoLaparo~\citep{wang2022autolaparo} & $1$ & $216\times384$ & $7$ & $21$ \\
CATARACTS~\citep{zisimopoulos2018deepphase} & $5$ & $216\times384$ & $19$ & $50$ \\
GTEA~\citep{fathi2011learning} & $15$ & $216\times288$ & $11$ & $28$ \\
50Salads~\citep{stein2013combining} & $5$ & $216\times288$ & $19$ & $50$ \\
\hline
\end{tabular}
}
\end{center}
\end{table}

\section{Metrics} \label{sec:metrics}

\textbf{Mean video-wise accuracy:}
We compute the mean video-wise accuracy, i.e. we compute one accuracy score per video and average over videos.   

\textbf{Mean video-wise balanced accuracy:}
We compute one balanced accuracy score per video and average over videos. Undefined cases do not occur here.

\textbf{Precision, Recall, Jaccard, F1:}
Per class, we compute true positives, false positives and false negatives over the entire test set. Then a single score is computer per class and averaged over classes. Only precision can encounter undefined values in this case, which we set to zero.

\textbf{Relaxed metrics for SOTA comparison:}
Some methods (including the current state-of-the-art methods Trans-SVNet~\citep{gao2021trans} and TMRNet~\citep{jin2021temporal}) report relaxed versions of metrics on the CHolec80 dataset, which were proposed in the \emph{M2CAI 2016 Challenge}~\citep{stauder2016tum}. Here, errors around phase transitions are ignored due to their ambiguity. Implementation details can be found in TMRNet's code\footnote{\url{https://github.com/YuemingJin/TMRNet/blob/main/code/eval/result/matlab-eval/}}.

\textbf{Weighted Mean Absolute Error (wMAE):}
We follow the original instrument anticipation approach~\citep{rivoir2020rethinking} and report \emph{wMAE} as our main metric. This is the mean of the frame-wise mean absolute errors inside \emph{(inMAE)} and outside \emph{(outMAE)} of the horizon. Specifically for each instrument, predicitons on all test videos are concatenated into a single sequence, to compute single \emph{inMAE} and \emph{outMAE} (and thus \emph{wMAE}) scores per instrument, which are then averaged.

\end{document}